\documentclass[a4paper,fleqn]{cas-dc}

\usepackage[authoryear,longnamesfirst]{natbib}

\usepackage{amsmath,amsfonts}
\usepackage{algorithmic}
\usepackage{array}
\usepackage[caption=false,font=normalsize,labelfont=sf,textfont=sf]{subfig}
\usepackage{textcomp}
\usepackage{stfloats}
\usepackage{url}
\usepackage{verbatim}
\usepackage{graphicx}

\usepackage{multicol}
\usepackage{tabularray}
\usepackage{array}

\usepackage{pdflscape}
\usepackage{tabularx}
\usepackage{multirow}
\usepackage{booktabs}
\usepackage{wrapfig}
\usepackage{lineno}
\usepackage{framed}
\usepackage{caption}
\usepackage{subcaption}
\usepackage{fancyhdr}
\usepackage{hyperref}
\usepackage{amsmath} 
\usepackage{cleveref}

\usepackage{todonotes}
\usepackage{booktabs}

\newboolean{showcomments}
\setboolean{showcomments}{true} 
\ifthenelse{\boolean{showcomments}}
{\newcommand{\nb}[2]{
  \fcolorbox{black}{yellow}{\bfseries\sffamily\scriptsize#1}
  {\sf$\blacktriangleright$\textit{#2}$\blacktriangleleft$}
 }
 
}
{\newcommand{\nb}[2]{}
 
}

\def\tsc#1{\csdef{#1}{\textsc{\lowercase{#1}}\xspace}}
\tsc{WGM}
\tsc{QE}
\tsc{EP}
\tsc{PMS}
\tsc{BEC}
\tsc{DE}

\begin{document}
\let\WriteBookmarks\relax
\def\floatpagepagefraction{1}
\def\textpagefraction{.001}

\shorttitle{Scraping the Shadows}

\shortauthors{CV Radhakrishnan et~al.}

\title [mode = title]{Scraping the Shadows: Deep Learning Breakthroughs in Dark Web Intelligence}                      



%
\author[1]{Ingmar Bakermans}



\ead{i.c.bakermans@gmail.com}


\credit{Conceptualization of this study, Methodology, Software}

\affiliation[1]{organization={Jheronimus Academy of Data Science},
    city={'s-Hertogenbosch},
    country={The Netherlands}}

\author[1]{Daniel De Pascale}
\ead{d.de.pascale@tue.nl}

 \author[2]{Gonçalo Marcelino}
\ead{g.barretoferreiramarcelino@uva.nl}


\affiliation[2]{organization={University of Amsterdam},
    city={Amsterdam},
    country={The Netherlands}}

\author%
[3]{Giuseppe Cascavilla}
\cormark[1]
\ead{g.cascavilla@tilburguniversity.edu}

\affiliation[3]{organization={Tilburg University},
    city={Tilburg},
    country={The Netherlands}}

\cortext[cor1]{Corresponding author}

\author%
[2]{Zeno Geradts}
\ead{z.j.m.h.geradts@uva.nl}




\begin{abstract}
Darknet markets (DNMs) facilitate the trade of illegal goods on a global scale. Gathering data on DNMs is critical to ensuring law enforcement agencies can effectively combat crime. Manually extracting data from DNMs is an error-prone and time-consuming task. Aiming to automate this process we develop a framework for extracting data from DNMs and evaluate the application of three state-of-the-art Named Entity Recognition (NER) models, ELMo-BiLSTM \citep{ShahEtAl2022}, UniversalNER \citep{ZhouEtAl2024}, and GLiNER \citep{ZaratianaEtAl2023}, at the task of extracting complex entities from DNM product listing pages. We propose a new annotated dataset, which we use to train, fine-tune, and evaluate the models. Our findings show that state-of-the-art NER models perform well in information extraction from DNMs, achieving 91\% Precision, 96\% Recall, and an F1 score
of 94\%. In addition, fine-tuning enhances model performance, with UniversalNER achieving the best performance. 
\end{abstract}




\maketitle

\section{Introduction}

The Internet connects global societies, enabling worldwide communication. It consists of three layers: the surface web, deep web, and dark web. The surface web is publicly accessible and indexed by search engines like Google and Bing \citep{GuptaMaynardAhmad2019}. The deep web contains restricted content such as databases and paywalled sites, requiring credentials for access \citep{NgoMarcumBelshaw2023}. Finally, the dark web, with its inherent anonymity, has become a hub for illicit activities, including drug trafficking, the trade of stolen financial information, and identity theft. Cybercriminals exploit this anonymity, collaborate in criminal forums, and recruit new members, all while evading detection \citep{ColeLatifChowdhury2021}. The lack of oversight within the dark web facilitates a significant volume of unreported and undetected cybercrimes. Darknet markets (DNMs), which function as e-commerce platforms for illegal goods, are central to this criminal ecosystem. Since the emergence of the first large-scale market, Silk Road, in 2011, DNMs have proliferated, with illegal drug sales being a primary driver of activity \citep{MeehanFarmer2023}. These markets present a substantial challenge for law enforcement agencies (LEAs), as they serve as vital intelligence sources but are difficult to monitor.

The dark web poses significant challenges for data collection due to its anonymity, security measures, and the dynamic nature of its content. Traditional web scraping techniques are often ineffective in this environment, struggling with issues like CAPTCHA protections, unstructured data, and the complexities of authentication and cookie management. These obstacles make it difficult for LEAs to gather the information necessary to combat illicit activities on DNMs
\citep{crator_crawler,10558720,CASCAVILLA2021102258}. 

To overcome these challenges, this study investigates advanced automated web scraping techniques that leverage Natural Language Processing (NLP) and machine learning with a particular focus on Named Entity Recognition (NER). NER approaches can identify and classify entities within unstructured text data, enabling more precise extraction of relevant information from DNMs. By integrating NER approaches and other NLP techniques into the web scraping pipeline, we aim to improve the efficiency and effectiveness of data collection. Our study evaluates state-of-the-art scraping methodologies, developing proofs of concept to test their scalability, robustness, and adaptability in the complex environment of the dark web.

To achieve this goal, this research defines the following research question (RQ): 

\begin{center}
    \textbf{RQ:} To what extent does our proof-of-concept automated scraping approach for DNMs compare to state-of-the-art techniques? 
\end{center}

This question arises from the observation made by Khder et al. \citep{Khder2021} that NLP can enhance web scraping processes, particularly in the areas of data cleaning, filtering, and organization. The goal is to assess how well the proposed approach performs in relation to established methods.
Furthermore, the study by Pichiyan et al. \citep{PichiyanEtAl2023} emphasizes the potential of integrating
NLP techniques into the web scraping pipeline, to extract and
analyze unstructured text data effectively and efficiently.

To address the main research question, we use a twofold approach. First, we evaluate each method individually by analyzing its performance in existing literature. Then, we apply these approaches to our dataset without training, using a zero-shot classification method \citep{ChangEtAl2008}. We also develop a proof of concept for each approach and compare their results to assess applicability, scalability, adaptation, and flexibility. This approach is examined through the following sub-research questions (sRQ):

\begin{center}
    \textbf{sRQ$1$:} To what extent can our proposed NER approaches scrape data from DNMs using zero-shot and fine-tuning methods?
\end{center}

\begin{center}
    \textbf{sRQ$2$:} Which proof of concept exhibits the best performance in automated dark web scraping with regards to recall, precision, and F1?
\end{center}

Evaluating the robustness of the proofs of concept is critical. In this context, we define \textit{robustness} as the method's ability to maintain high performance across diverse web environments, withstand common dark web challenges such as dynamic content, anti-scraping mechanisms, variations in data structure, and adaptability to different web technologies and standards. To assess the robustness of the chosen methods, we adopt the technique introduced by Guo et al. \citep{GuoEtAl2023}. This approach provides insights into the reliability, durability, and potential for long-term implementation of the chosen method in varied web scraping scenarios. Therefore, we describe our third research question as: 

\begin{center}
    \textbf{sRQ$3$:} In comparison with existing studies, how robust are the proposed approaches for automated dark web scraping?
\end{center}
This work makes several critical contributions to the fields of cybersecurity and digital forensics, specifically in the analysis of Darknet Markets:

\begin{itemize}
  \item \textbf{Comprehensive comparison of NER tagging models:}  This study systematically compares existing Named Entity Recognition (NER) tagging models for automated DNM scraping. This focus on methodologies rather than tools aims to identify the most effective and accurate methods to retrieve information from DNMs.
  \item \textbf{Bridging the gap for LEAs:} Our proposed approach addresses a gap in the literature regarding automated web scraping methods for LEAs. By evaluating methods through the lens of practical law enforcement requirements, this study validates new approaches aimed at combating cybercrime on DNMs.
  \item \textbf{Solution to data scarcity in DNM research:} Our study highlights a data scarcity problem in this research field and proposes an automated, scalable data collection framework to collect and analyze the increasing volume of data on DNMs. This framework is crucial for timely and effective law enforcement interventions
  \item \textbf{Dataset for future research:} As part of this study, we compiled and annotated a new dataset of data scraped from DNM pages. This dataset facilitates the comparison of scraping methods and serves as a benchmark for developing more sophisticated data extraction and analysis tools tailored to the needs of LEAs.
\end{itemize}


Preliminary research suggests that integrating advanced machine learning and natural language processing (NLP) techniques could substantially improve the efficiency and accuracy of data extraction from DNMs. Utilizing dynamic cookies and intelligent crawling mechanisms shows potential for overcoming the security barriers that traditionally hinder web scraping on Darknet markets. Early findings indicate that combining these advanced approaches with conventional web scraping techniques yields a more effective and comprehensive strategy for collecting data from DNMs, thereby enhancing the capability to monitor and analyze illicit activities on the dark web.

\section{Related Work}
The rise of DNMs following the launch of Silk Road in 2011 has sparked significant academic interest, leading to various studies on these platforms \citep{MeehanFarmer2023}. Yannikos et al. \citep{YannikosHeegerBrockmeyer2019} introduced a framework to analyze product prices and supply on DNMs, featuring an automated pipeline for data collection, analysis, and correlation with external sources, such as news websites. Similarly, De Pascale et al. \citep{DePascaleEtAl} developed a platform for LEAs to analyze large-scale data from the Dark Web \citep{DePascaleEtAl}, employing an intelligent crawler to gather information from three DNMs: Agartha, Cannazon, and Dark Market. The data is collected using a crawler designed to navigate the complex structure of the dark web. The crawler can bypass basic security layers, such as login form pages with simple CAPTCHAs. Moreover, it allows for integrating multiple CAPTCHAs with real-time manual intervention by the user \citep{crator_crawler}. 


These studies underscore the critical role of data retrieval from DNMs in the existing literature. Kalpakis et al. \citep{KalpakisEtAl2017} emphasize the Dark Web's anonymity as a catalyst for illegal activities, making it a focal point for LEAs interested in gathering Open Source Intelligence (OSINT). The need for advanced technologies to facilitate data collection and identify criminal activities on the Dark Web is further highlighted by Alquwayzani et al. \citep{AlquwayzaniEtAl2023}, who argue that artificial intelligence and machine learning can significantly enhance OSINT efforts. Szymoniak and Foks \citep{SzymoniakFoks2024} corroborate these findings, advocating for the use of dark web data in LEA monitoring efforts, while also noting the challenges of technical proficiency required to navigate these spaces safely.

Despite the progress, the literature lacks comprehensive comparisons of automated scraping methods, focusing instead on tool comparisons as noted by Uzun \citep{Uzun2020}. Furthermore, the applicability of advanced information extraction models in the context of DNMs remains underexplored. Among the existing models, the ELMO-BiLSTM-CNN proposed by Shah et al. \citep{ShahEtAl2022} stands out, outperforming its predecessor with a Precision of 97.34\%, Recall of 97.60\%, and F1-score of 96.13\%. However, our analysis of several studies, including those by Kejriwal and Szekely \citep{KejriwalSzekely2017}, Gur et al. \citep{GurEtAl2023}, and Yannikos et al. \citep{YannikosHeegerBrockmeyer2019}, led us to adopt the UniversalNER \citep{ZhouEtAl2024} and GLiNER \citep{ZaratianaEtAl2023} models for their state-of-the-art zero-shot capabilities. Notably, these models have yet to be tested on dark web datasets, presenting further research opportunities.

Additionally, there is a pressing need for more reliable and representative DNM data. Ball et al. \citep{BallBroadhurst2021} highlight the challenges of manual data collection from DNMs, which is labor-intensive and inadequate for the growing scale of these markets. They emphasize the necessity for automated web scraping and classification methods to effectively manage the increasing volume of DNM data.
\section{Experimental Design}


In this Section we describe the methodologies employed in the comparison of the different NER tagging models in the context of DNMs scraping.

\subsection{Data Gathering}
\begin{figure}[h!]
    \centering
   \includegraphics[width=0.5\textwidth]{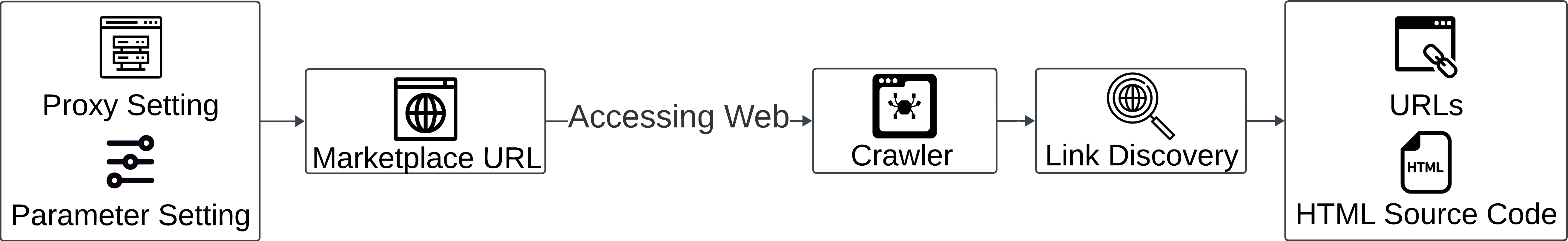}
    \caption{Crawler infrastructure introduced by Shah et al. \citep{ShahEtAl2022}.}
    \label{fig:Crawler-infra}
\end{figure}

The dataset we use in our research includes data from Pascale et al. \citep{DePascaleEtAl} Agartha and Dark Market and incorporates unpublished DNM data from four other DNMs, namely 
Berlusconi Market, Cannahome, Silkroad, and Palmetto market. Furthermore, we crawled new data from Cocorico Market using an intelligent dark web crawler presented in Figure \ref{fig:Crawler-infra}. The crawler starts its process at the seed URL. In our approach, the seed URL is the product overview page of the Cocorico market. This overview page includes URLs to product pages, which are present in the overview. These URLs are collected, and the crawler downloads the underlying HTML page. This process is conducted three times.
If a URL appears repeatedly, the duplicate page is not downloaded again. This approach ensures we do not duplicate entries in our crawled dataset. To prevent the web crawler from endlessly crawling, we set a maximum limit of one million stored links and a maximum crawling time of 86,400 seconds (equivalent to 24 hours). Due to the common occurrence of DDoS attacks on DNM pages, these markets often employ preventative measures such as IP-blockers that activate when specific IPs send many requests in a short period. We have implemented socksh5 proxies, random wait timers, and delays in sending requests to address this issue.

The data from Agartha, Dark Market, Berlusconi Market, Cannahome, Silkroad DNMs, and the crawled data from the Cocorico market are used to train the models. The data from the Palmetto Marketplace is used to test the robustness of the best-performing approach. Table \ref{tab:market_summary} presents the number of HTML documents and the average tokens per DNM page. The dataset contains two languages. The crawled webpages from Cocorico Market are in French; the rest consist of English text.

\begin{table}[hbt!]
\centering
\scalebox{0.8}{
\normalsize
\begin{tabular}{p{0.16\textwidth}p{0.16\textwidth}p{0.16\textwidth}}
\hline
\textbf{Market} & \textbf{\# of webpages} & \textbf{average \# tokens} \\ \hline
Argartha Market & 1334 & 230 \\ \hline
Berlusconi Market & 123 & 184 \\ \hline
Cocorico Market & 1936 & 153 \\ \hline
Dark Market & 658 & 175 \\ \hline
Silkroad & 1649 & 267 \\ \hline
Cannahome & 1037 & 1385 \\ \hline
\textbf{Total} & 6737 & 389 \\ \hline
\end{tabular}
}
\caption{Dataset summary representing the number of HTML documents and average number of tokens per DNM.}
\label{tab:market_summary}
\end{table}


\subsection{Data Preparation}


Given a product page from a DNMs, we use an HTML parser and filter out all the tags to retain only text and then convert multiple sequential white spaces into single spaces.  We use RegEx to identify and label the specific entities within the HTML files. We obtained relevant entities: product, market name, product price, category, vendor, stock, and product views. We created tailor-made RegEx patterns for each marketplace, as shown in Table \ref{tab:combined-regex-patterns1} and Table \ref{tab:combined-regex-patterns2}. 

\begin{table}[h!]
\centering
\scalebox{0.75}{
\normalsize
\begin{tabular}{|ll|}
\hline
\multicolumn{1}{|l|}{\textbf{Entity}}      & \textbf{RegEx Pattern}\\ \hline
\multicolumn{2}{|c|}{\textbf{Agartha Item}}                                                                                            \\ \hline
\multicolumn{1}{|l|}{market\_name}         & (\textbackslash w+)\textbackslash s+purchase                                              \\ \hline
\multicolumn{1}{|l|}{product}              & (?:listings.*?)1listings\textbackslash s+(.+?)\textbackslash s+message                    \\ \hline
\multicolumn{1}{|l|}{model}                & category\textbackslash s+(\textbackslash w+)                                              \\ \hline
\multicolumn{1}{|l|}{quantity\_in\_stock}  & availability (\textbackslash d+)                                                          \\ \hline
\multicolumn{1}{|l|}{product\_price}       & price (\textbackslash d+\textbackslash .\textbackslash d+)                                \\ \hline
\multicolumn{1}{|l|}{product\_description} & listings (.+?) purchase                                                                   \\ \hline
\multicolumn{1}{|l|}{product\_views}       & (\textbackslash d+)\textbackslash s+\textbackslash d+,\textbackslash d+\textbackslash s?€ \\ \hline
\multicolumn{1}{|l|}{vendor\_name}         & vendor\textbackslash s+(\textbackslash w+)                                                \\ \hline
\multicolumn{2}{|c|}{\textbf{Agartha Purchase}}                                                                                        \\ \hline
\multicolumn{1}{|l|}{market\_name}         & (\textbackslash w+)\textbackslash s+purchase                                              \\ \hline
\multicolumn{1}{|l|}{product}              & (?:purchase.*?)1purchase\textbackslash s+(.+?)\textbackslash s+category                   \\ \hline
\multicolumn{1}{|l|}{model}                & category\textbackslash s+(\textbackslash w+)                                              \\ \hline
\multicolumn{1}{|l|}{quantity\_in\_stock}  & availability (\textbackslash d+)                                                          \\ \hline
\multicolumn{1}{|l|}{product\_price}       & (\textbackslash d+\textbackslash .\textbackslash d+)\textbackslash s+(usd|btc)            \\ \hline
\multicolumn{1}{|l|}{product\_description} & listings (.+?) purchase                                                                   \\ \hline
\multicolumn{1}{|l|}{product\_views}       & (\textbackslash d+)\textbackslash s+\textbackslash d+,\textbackslash d+\textbackslash s?€ \\ \hline
\multicolumn{1}{|l|}{vendor\_name}         & vendor\textbackslash s+(\textbackslash w+)                                                \\ \hline
\multicolumn{2}{|c|}{\textbf{Berlusconi}}                                                                                              \\ \hline
\multicolumn{1}{|l|}{market\_name}         & \textbackslash berlusconi\textbackslash b                                                 \\ \hline
\multicolumn{1}{|l|}{product} & $^(.+?)$\textbackslash s\d+(?:\textbackslash .\textbackslash d+)?\textbackslash s*eur \\ \hline
\multicolumn{1}{|l|}{model}                & class\textbackslash s+(\textbackslash w+)                                                 \\ \hline
\multicolumn{1}{|l|}{quantity\_in\_stock}  & (\textbackslash d+) s+in stock                                                            \\ \hline
\multicolumn{1}{|l|}{product\_price}       & (\textbackslash d+(?:\textbackslash .\textbackslash d+)?)(?=\textbackslash s+eur)         \\ \hline
\multicolumn{1}{|l|}{product\_description} & avis (.+?) modèle                                                                         \\ \hline
\multicolumn{1}{|l|}{product\_views}       & (\textbackslash d+)\textbackslash s+\textbackslash d+,\textbackslash d+\textbackslash s?€ \\ \hline
\multicolumn{1}{|l|}{vendor\_name}         & vendor\textbackslash s+(\textbackslash w+)                                                \\ \hline
\multicolumn{2}{|c|}{\textbf{Cannahome}}                                                                                               \\ \hline
\multicolumn{1}{|l|}{market\_name}         & (\textbackslash w+)\textbackslash s+purchase                                              \\ \hline
\multicolumn{1}{|l|}{product}              & details (.+?) availability                                                                \\ \hline
\multicolumn{1}{|l|}{model}                & category\textbackslash s+(\textbackslash w+)                                              \\ \hline
\multicolumn{1}{|l|}{quantity\_in\_stock}  & availability (\textbackslash d+)                                                          \\ \hline
\multicolumn{1}{|l|}{product\_price}       & escrow (\textbackslash d+.\textbackslash d+)                                              \\ \hline
\multicolumn{1}{|l|}{product\_description} & avis (.+?) modèle                                                                         \\ \hline
\multicolumn{1}{|l|}{product\_views}       & (\textbackslash w+)\textbackslash s+orders                                                \\ \hline
\multicolumn{1}{|l|}{vendor\_name}         & vendor\textbackslash s+(\textbackslash w+)                                                \\ \hline
\end{tabular}
}
\caption{RegEx Patterns for retrieving entities from various DNMs.}
\label{tab:combined-regex-patterns1}
\end{table}

\begin{table}[h!]
\centering
\scalebox{0.8}{
\normalsize
\begin{tabular}{|ll|}
\hline
\multicolumn{1}{|l|}{\textbf{Entity}}      & \textbf{RegEx Pattern}\\ \hline
\multicolumn{2}{|c|}{\textbf{Cocorico}}                                                                                                \\ \hline
\multicolumn{1}{|l|}{market\_name}         & \textbackslash bcocorico\textbackslash s market\textbackslash b                           \\ \hline
\multicolumn{1}{|l|}{product}              & recherche (.+?) description                                                               \\ \hline
\multicolumn{1}{|l|}{model}                & modèle\textbackslash s+(.+?)\textbackslash s+disponibilité                                \\ \hline
\multicolumn{1}{|l|}{quantity\_in\_stock}  & disponibilité :(\textbackslash d+)                                                        \\ \hline
\multicolumn{1}{|l|}{product\_price}       & (\textbackslash d+, d+)\textbackslash s?€                                                 \\ \hline
\multicolumn{1}{|l|}{product\_description} & avis (.+?) modèle                                                                         \\ \hline
\multicolumn{1}{|l|}{product\_views}       & (\textbackslash d+)\textbackslash s+\textbackslash d+,\textbackslash d+\textbackslash s?€ \\ \hline
\multicolumn{1}{|l|}{vendor\_name}         & (\textbackslash b\textbackslash w+\textbackslash b)\textbackslash s+rating                \\ \hline
\multicolumn{2}{|c|}{\textbf{Dark Market}}                                                                                             \\ \hline
\multicolumn{1}{|l|}{market\_name}         & (\textbackslash w+)\textbackslash s+purchase                                              \\ \hline
\multicolumn{1}{|l|}{product}              & 1 (.+?) quality                                                                           \\ \hline
\multicolumn{1}{|l|}{model}                & type\textbackslash s+(\textbackslash w+)                                                  \\ \hline
\multicolumn{1}{|l|}{quantity\_in\_stock}  & leftsold (\textbackslash d+)                                                              \\ \hline
\multicolumn{1}{|l|}{product\_price}       & offers (\textbackslash d+\textbackslash .+\textbackslash d+)                              \\ \hline
\multicolumn{1}{|l|}{product\_description} & listings (.+?) quality                                                                    \\ \hline
\multicolumn{1}{|l|}{product\_views}       & (\textbackslash d+)\textbackslash s+\textbackslash d+,\textbackslash d+\textbackslash s?€ \\ \hline
\multicolumn{1}{|l|}{vendor\_name}         & information\textbackslash s+(\textbackslash w+)                                           \\ \hline
\multicolumn{2}{|c|}{\textbf{Silkroad}}                                                                                                         \\ \hline
\multicolumn{1}{|l|}{market\_name}         & \textbackslash bsilk\textbackslash s road\textbackslash b                                 \\ \hline
\multicolumn{1}{|l|}{product}              & usd+(.+?)\textbackslash s+price                                                           \\ \hline
\multicolumn{1}{|l|}{model}                & category+(.+?)\textbackslash s+stock                                                      \\ \hline
\multicolumn{1}{|l|}{quantity\_in\_stock}  & remaining\textbackslash s*(\textbackslash d+)                                             \\ \hline
\multicolumn{1}{|l|}{product\_price}       & price s*(\textbackslash d+)                                                               \\ \hline
\multicolumn{1}{|l|}{product\_description} & avis (.+?) modèle                                                                         \\ \hline
\multicolumn{1}{|l|}{product\_views}       & (\textbackslash d+)\textbackslash s+\textbackslash d+,\textbackslash d+\textbackslash s?€ \\ \hline
\multicolumn{1}{|l|}{vendor\_name}         & listings\textbackslash s+(\textbackslash w+)                                              \\ \hline
\end{tabular}
}
\caption{Combined RegEx patterns for retrieving entities from various DNMs.}
\label{tab:combined-regex-patterns2}
\end{table}

We manually checked the retrieved entities to ensure the labeled data was correct and sufficient for model training and that the labeling accuracy was above 90\%. String normalization was crucial in refining the data by eliminating irrelevant keywords, symbols, and stopwords (i.e., \&, *, and ; , :). 

\subsection{Modeling} 
This section analyzes the models employed for NER. The models we utilize are ELMo-embedded BiLSTM model \citep{ShahEtAl2022}, UniversalNER \citep{ZhouEtAl2024}, and GLiNER \citep{ZaratianaEtAl2023}. 

\subsubsection{Bidirectional-LSTM}
\begin{figure}[hbt!]
    \centering
    \includegraphics[width=0.5\textwidth]{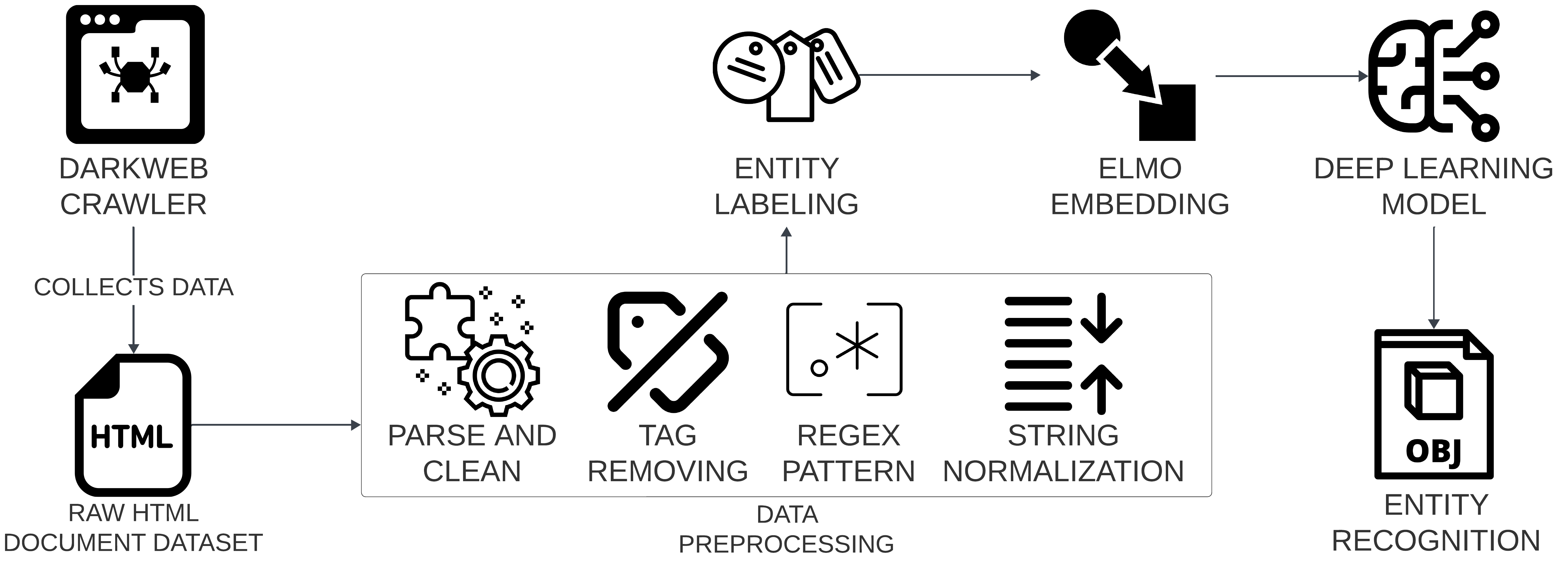}
    \caption{ELMo-BiLSTM-CNN model pipeline.}
    \label{fig:ELMO-Embedding}
\end{figure}

The Named Entity Recognition using a Bidirectional-LSTM approach is based on the work of Shah et al. \citep{ShahEtAl2022}. The authors propose an ELMo-BiLSTM-CNN model for detecting dark web e-commerce entities. This method serves as an effective data extraction approach and a strong baseline for comparing zero-shot NER models like UniversalNER and GLiNER. In comparison to Shah et al., our model can retrieve specific entities from the dark web page instead of all entities with specific labels and is applicable in the domain of multiple languages. Our proposed approach consists of three layers: an ELMO embedding layer, a Bi-LSTM layer, and a convolutional neural network (CNN) layer. Since the ELMo-embedding system used by Shah et al. \citep{ShahEtAl2022} and introduced by Peeters et al. \citep{DBLP:journals/corr/abs-1802-05365} is not suitable in the context of multi-language, we employed the multi-language ELMo-embedding system introduced by Che et al. \citep{CheEtAl2018}. A schematic presentation of the development steps taken to create our model is shown in Figure \ref{fig:ELMO-Embedding}.

The multi-language ELMo-embedding system \citep{CheEtAl2018} employed in this work is trained on a set of 20-million-word data randomly sampled from the raw text released by the shared task (wiki dump + common crawl) for each language included in the system. The multi-language ELMo-embedding system in this study is trained on 20 million words sampled from shared task data (wiki dump + common crawl) for each language. Word embeddings map words to vectors, linking human understanding to machine processing \citep{BirundaDevi2021}. ELMo embeddings are contextual, with representations changing dynamically based on the surrounding text. This approach enables entity extraction from HTML source code. 

In our approach, the input sequence for the model consists of an ELMo-embedding for all tokens in the sentence, with the order staying intact. We add the ELMo-embedding to a two-layer biLSTM model with character convolution one-dimensional layers. 
Our proposed approach uses a Bi-LSTM to retrieve defined entities from the input sequence. 
The sigmoid function utilized by the forget gate \(f_t\) is defined in the same way as done in the research conducted by Shah et al. \citep{ShahEtAl2022}.

The final layer of our model is a Convolutional One-Dimensional (Conv1D) layer, as described by Xie et al. \citep{XieEtAl2024}, optimized for multi-class legal citation text classification. Conv1D layers are well-suited for sequential data like text, making them ideal for this task \citep{XieEtAl2024}. In our architecture, the Conv1D layer is combined with an AveragePooling layer, a design choice that mitigates overfitting, as evidenced by comparisons with Shah et al. \citep{ShahEtAl2022}. The Conv1D layer plays a critical role in extracting entity-related features from HTML data.


\subsubsection{UniversalNER}
The second approach is the UniversalNER method introduced by Zhou et al. \citep{ZhouEtAl2024}. UniversalNER is an instruction-tuned language model obtained through targeted distillation from large language models (LLMs). 

\begin{figure}
    \centering
    \includegraphics[width=0.35\textwidth]{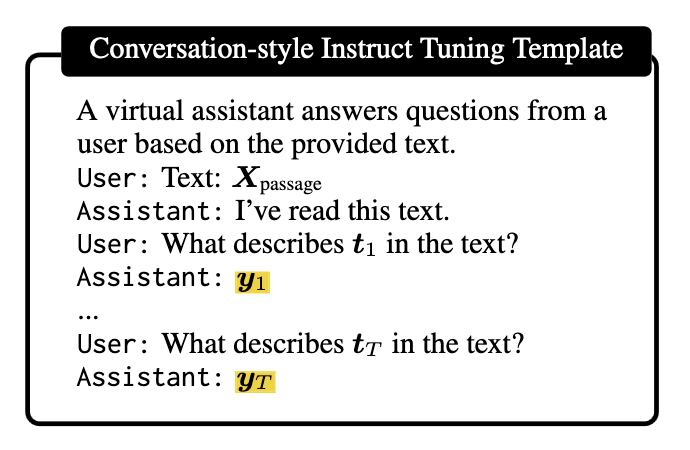}
    \caption{The conversation-style template. The conversation is used to tune language models. Only the
    highlighted parts are used to compute the loss \citep{ZhouEtAl2024}.}
    \label{fig:UniversalNER}
\end{figure}

The authors obtained the final model by training student models to replicate the capabilities of the original LLM for the specific application of open NER. The training uses data construction and instruction tuning to distill the knowledge from ChatGPT into smaller, more cost-efficient models such as the UniversalNER model, with the data construction Zhou et al. \citep{ZhouEtAl2024} sampled inputs from the Pile corpus, which compiles 22 distinct English sub-datasets. They chunked the articles from the Pile corpus into passages reaching a maximum length of 256 tokens and randomly sampled 50K passages as the inputs. Zhou et al. used  ChatGPT (gpt-3.5-turbo-0301) to generate the entity mentions and their associated types on the sampled passages. With this prepared data, Zhou et al. trained the Llama model introduced by Touvron et al. \citep{TouvronEtAl2023}, fine-tuning the Llama model on the task of NER.
To ensure stability, the authors set the generation temperature to 0, which means the model will always select the word with the highest probability for a particular entity. After filtering out unparseable outputs and inappropriate entities, the fine-tuning resulted in 13,020 distinct entity types. The distribution of the extracted entity types showed a heavy tail where the top 1\% accounted for 74\% of the total frequencies. With instruction tuning, we provide the model with prompts and change the model's parameters to improve the performance of the goal described by the prompts. Figure \ref{fig:UniversalNER} shows the instruction tuning template and represents a conversation-style tuning format. This approach presents a text \(X_{passage}\) as input to the UniversalNER model. The entity type we want to retrieve from the text is defined as \(t_i\) and formatted as ``What describes \(t_i\) in the text?''. In addition, the Language Model (LM) is tuned to generate a structured output \(y_i\) formatted in a JSON list containing all tokens labeled as entity \(t_i\) within the text \(X_{passage}\). In the study by Zhou et al. \citep{ZhouEtAl2024}, the authors introduced a method for improving the model performance using supervised data. They remark on the risk of discrepancies in label definitions among datasets when using different datasets. This issue is addressed using dataset-specific instruction tuning templates that harmonize label definition discrepancies. The dataset we used in this study does not contain discrepancies in label definitions because the entities are labeled with rigid boundaries, meaning there is only one correct entity. Zhou et al. compared zero-shot models on datasets from different domains
the distilled models (UniversalNER) achieve better results than ChatGPT on all evaluated domains. In our experiment, we fine-tuned the UniversalNER model for dark web entity extraction. Due to out-of-memory (OOM) issues, we propose a new fine-tuning approach for UniversalNER.

\subsubsection{GLiNER}
The third approach we test is GLiNER \citep{ZaratianaEtAl2023} (Generalist Model for Named Entity Recognition using Bidirectional Transformer), a state-of-the-art machine learning model primarily developed for Named Entity Recognition. The GLiNER model is a derivation of UniversalNER. 
However, the UniversalNER approach has its limitations \citep{ZaratianaEtAl2023}: 

\begin{itemize}
  \item UniversalNER utilizes auto-regressive language models, which may be slow because they generate text one token at a time (token-by-token generation)
  \item UniversalNER relies on large models with billions of parameters, making them difficult to deploy in environments with limited computing resources.
\end{itemize}

By introducing GLiNER, Zaratiana et al. \citep{ZaratianaEtAl2023} address these limitations. The heavy auto-regressive language models from UniversalNER are changed in lighter bidirectional language models (BiLM), such as BERT, introduced by Devlin et al. \citep{DevlinEtAl2019} to address these limitations. The core concept of the GLiNER model is: ``treating the task of Open NER as matching entity type embedding to textual span representations in latent space, rather than as a generation task'' \citep{ZaratianaEtAl2023}. Following this belief, the approach naturally solves the scalability issues of auto-regressive models and allows for bidirectional context processing, which enables richer representation. In addition, the robustness of the GLiNER model enables handling languages that were not present in the training data. In the research conducted by Zaratiana et al. \citep{ZaratianaEtAl2023}, the authors included comparisons between ChatGTP, UniversalNER-7B, and GLiNER-L in a zero-shot and fine-tuned state. In all cases, UniversalNER-7B and GLiNER-L performed better than ChatGTP.

The GLiNER model employs a BiLM and takes as input entity type prompts and a sentence/text. In this way, GliNER can be more resource-efficient than the earlier introduced UniversalNER. Each entity is separated by a learned token [ENT]. The BiLM outputs representations for each token. Entity embeddings are passed into a FeedForward Network, while input word representations are passed into a span representation layer to compute embeddings for each span.  Finally, the model computes a matching score between entity and span representations (using dot product and sigmoid activation).

\subsection{Evaluation} 
We follow two approaches during model evaluation. First, we assess the performance of the models based on the F-score, Precision, and Recall. 

%




As a second step, we evaluate the robustness of the best-performing model. The robustness encompasses the model's ability to maintain high performance across diverse web environments, such as DNMs. Robustness is an essential measure due to the characteristics of the dark web, such as dynamic content, anti-scraping mechanisms, and variations in data structure. To assess robustness, we will adopt the method introduced by Guo et al. \citep{GuoEtAl2023}, tailored for evaluating deep learning models utilized in this study. This approach provides insights into the reliability, durability, and potential for long-term implementation of the chosen method in varied web scraping scenarios.

\section{Experiment}
\label{experiments}


\subsection{Data Understanding}

\begin{figure}[h!]
\centering
\begin{minipage}{0.5\textwidth}
  \centering
  \includegraphics[width=1.0\linewidth]{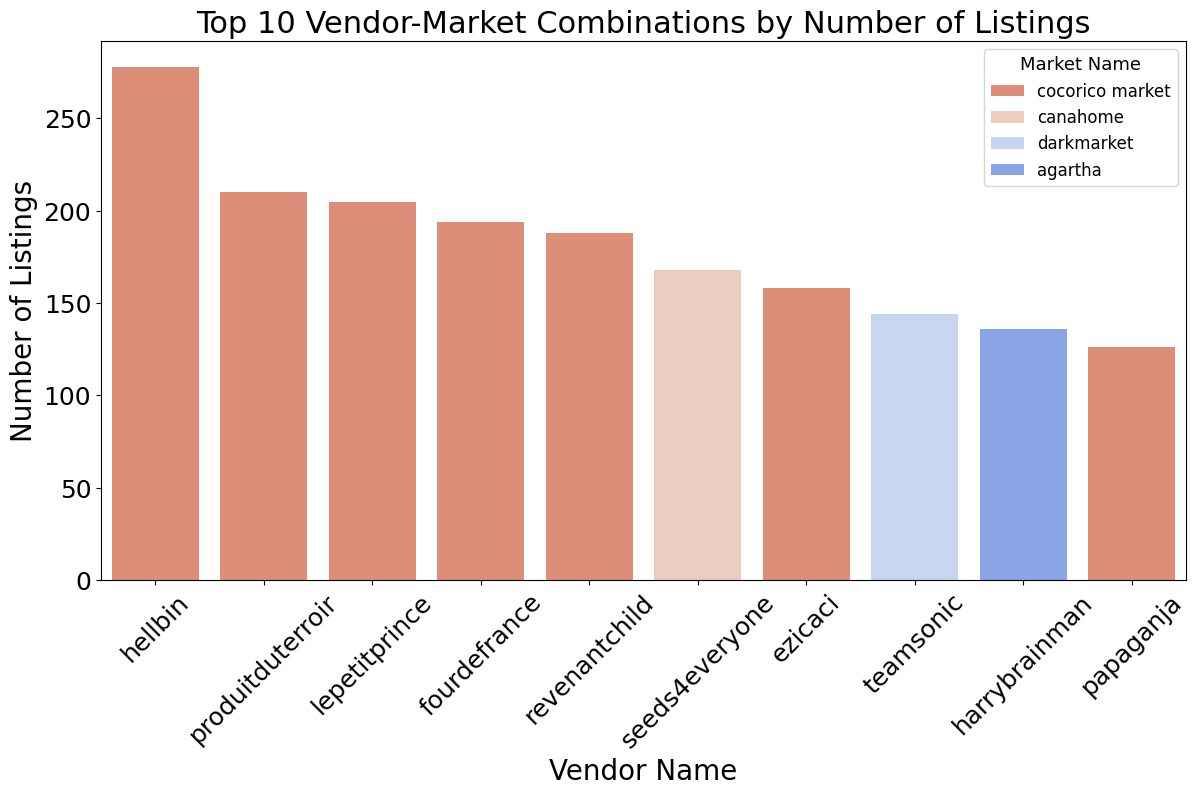}
  \captionof{figure}{Top 10 Vendor-DNM combinations.}
  \label{fig:vendor-market-combinations}
\end{minipage}\hfill
\begin{minipage}{0.5\textwidth}
  \centering
  \includegraphics[width=1.0\linewidth]{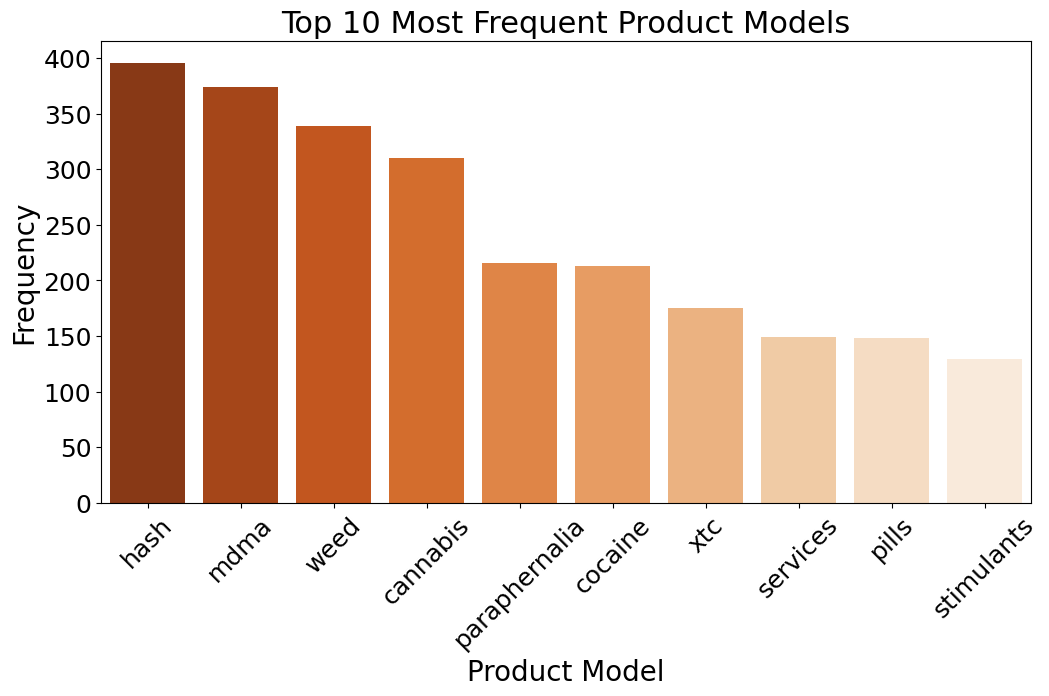}
  \captionof{figure}{Top 10 Models sold across all analyzed DNMs.}
  \label{fig:frequent-models}
\end{minipage}\hfill
\begin{minipage}{0.5\textwidth}
  \centering
  \includegraphics[width=1.0\linewidth]{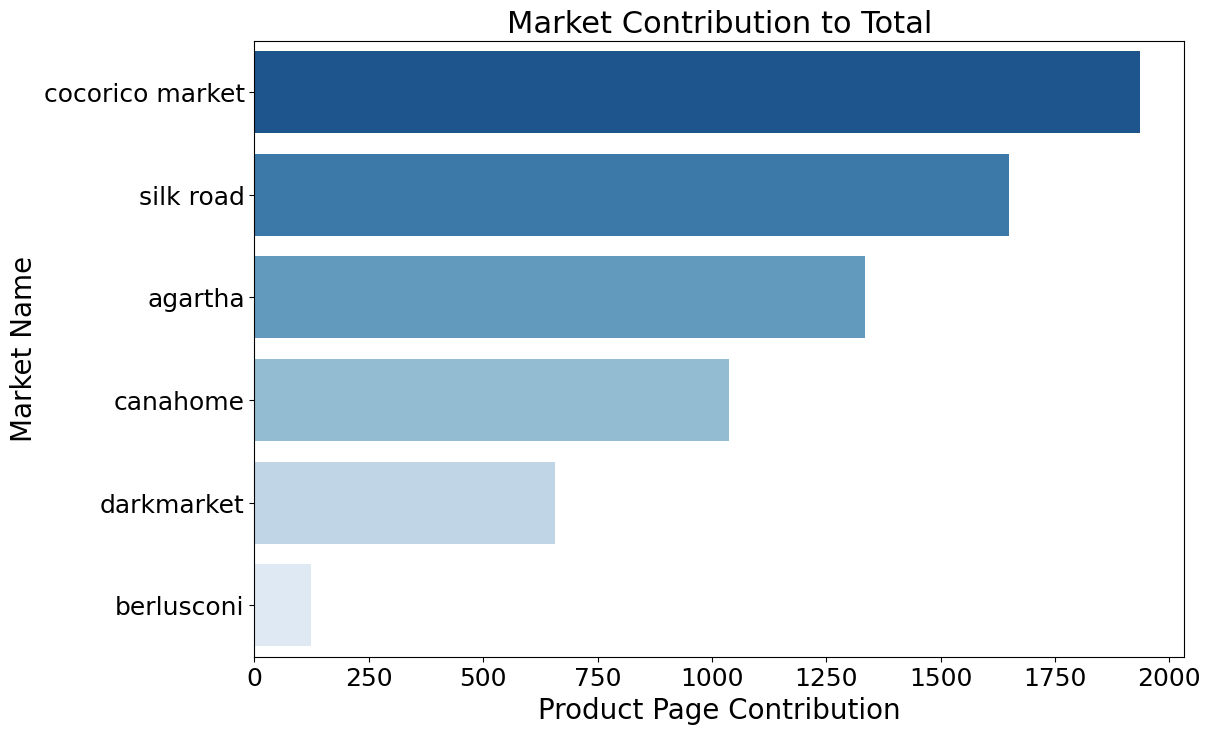}
  \captionof{figure}{Contribution on dataset per DNM.}
     \label{fig:total-listings}
\end{minipage}\hfill
\hfill
 \label{fig:dnmslistings-summary}
\end{figure}

We parsed the HTML pages using BeautifulSoup to extract and clean the data by removing tags, normalizing text, and eliminating inconsistencies like extra spaces and misplaced symbols. Given the noisy and irregular nature of dark web HTML data, normalization was essential before entity retrieval to preserve data integrity.
We then used regular expression (RegEx) to label the dataset. 
The process revealed exciting insights from the dataset acquired. Figures \labelcref{fig:vendor-market-combinations,fig:frequent-models,fig:total-listings} present some exploratory results on the data in more detail. Figure \ref{fig:total-listings} shows the number of listings crawled per marketplace and the share of the total dataset per marketplace. An important fact is that the content from Cocorico Market contains French text, while the other marketplaces are English. Figure \ref{fig:frequent-models} and Figure \ref{fig:vendor-market-combinations} give more descriptive details about the dataset, such as the top 10 sold categories on the marketplaces. In addition, we included a plot that presents the top vendors concerning product listings and to which marketplace they belong; this data is presented in Figure \ref{fig:vendor-market-combinations}.

\subsection{Model Training}

\subsubsection{Bidirectional-LSTM}
As the first approach, we introduce the Bidirectional-LSTM modeling approach. 
As discussed in the previous section, the dataset includes normalized and cleaned HTML source code along with labeled tokens linked to corresponding entities. 
Given the presence of both French and English language in the data, we adopted the multilingual ELMo embeddings system by Che et al. \citep{CheEtAl2018}. 
This embedding system uses the same hyperparameter settings as Peters et al. \citep{DBLP:journals/corr/abs-1802-05365} for the BiLM and the character CNN. The parameters are trained on 20-million-word data randomly sampled from raw text released by the shared task (including wiki dump and common crawl). Because we used a Keras pipeline to create the model and the Multi-language ELMo-embedding model cannot be implemented in this pipeline due to dependency constraints, we had to adapt to the implementation method. 

As described, the model applied consisted of an ELMo-embedding combined with a BILSTM-CNN model. We tokenized the normalized sentences using the Keras preprocessing mechanism and indexed them to achieve this construction. The indexation gave each token a unique number (id). Afterward, each token was embedded using ELMo for multi-languages, which converted the token into a vector with embedding 1 x 1024, which is standard for ELMo-embedding. Using this vector, ELMo generates context-aware embeddings. The embedding uses character-level representations, so the model learns embeddings for words outside of vocabulary and can place such words in context. The embeddings of each token in the corpus were stored in an embeddings matrix with a 34598 x 1024 dimension (words in corpus x ELMo-embedding). After creating the embedding matrix, the normalized text is converted to sequences of token IDs representing the sentences in numerical format; the same is done for the identified entities. The last step for preprocessing the input data is that the id sequences are padded to 3000 tokens to equalize the length of the input sequences.

\begin{figure}[h]
    \centering
    \includegraphics[width=0.4\textwidth]{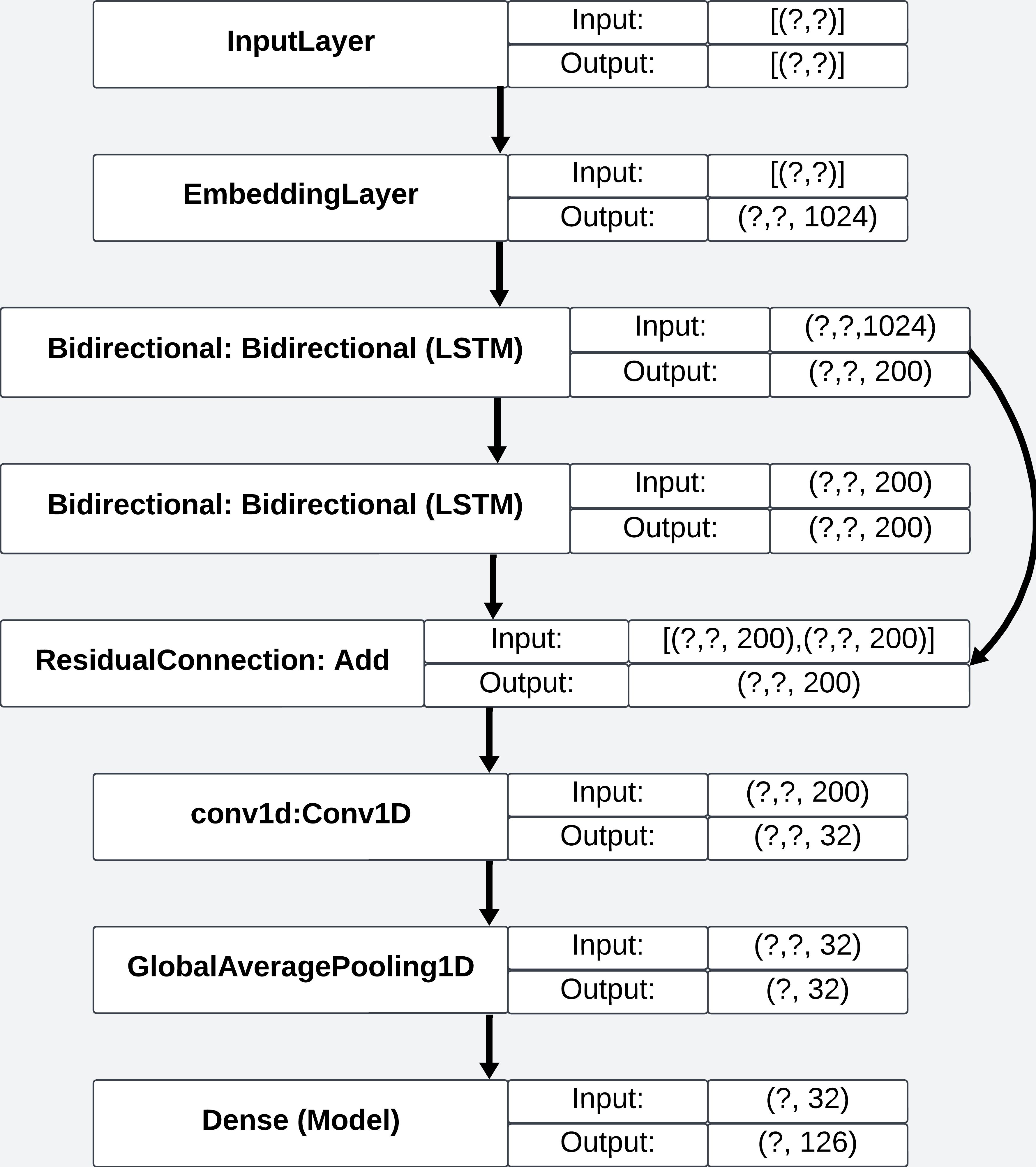}
    \caption{Design of our proposed ELMO-BiLSTM-CNN model.}
    \label{fig:Bi-LSTM}
\end{figure}

After preprocessing the data for the modeling phase, we designed the model. Figure \ref{fig:Bi-LSTM} presents a schematic presentation of the layers for designing and developing our deep-learning model:

\begin{itemize}
  \item \textbf{Token embedding:} In the Embedding Layer, the embedding matrix described is implemented. The embedding layers convert the sequences of token IDs into sequences of their connected ELMo-embeddings. Since the input text from the dark web can consist of long sequences of tokens, the padding size is 3000 tokens. 
  \item \textbf{Bi-LSTM:} For the Bi-LSTM layer, we followed the design proposed by Shah et al. \citep{ShahEtAl2022} where a bidirectional LSTM accepts a recurrent layer (e.g., the first LSTM layer). The embeddings created from the embedding layer are utilized in this layer. In addition, we used two Bi-LSTM layers with a residual link to the first Bi-LSTM. This residual link is shown in the Add Layer.
  \item \textbf{Conv1D:} After the residual Bi-LSTM layers a one-dimensional convolutional layer (Conv1D) is added to the model. This Conv1D layer, on top of the Bi-LSTM-CNN model, collects information using a convolutional process. After the information gathering, it utilizes this information as input to pass through the model from the model. The origin of the information gathered is the Add layer. Because the dimensions had to be reduced from 3 to 2 dimensional to satisfy the Dense layer requirements, we implemented an AveragePooling layer. This pivot, compared to \citep{ShahEtAl2022}, reduces the over-fitting of the model.
  \item \textbf{Dense:} We use a dense layer with the Softmax activation function in our model. The Softmax function is widely known for its effectiveness in managing multi-class classification issues. It serves as the activation function in the output layer of neural networks, calculating the multinomial probability for each class. In our neural network, Softmax is employed to address multi-class classification challenges, where it is necessary to determine class membership among more than two class labels. Our goal is to predict the multi-class label from the text, and the Softmax function provides the probability of each label, identifying the one with the highest probability as the predicted label.
\end{itemize}

We used the hyperparameters in Table \ref{tab:BiLSTM-hyperparameter-settings} to train the model. When training the model, we used the settings introduced by Shah et al. \citep{ShahEtAl2022}. 

\begin{table}[h!]
\centering
\scalebox{0.9}{
\begin{tabular}{p{0.2\textwidth}p{0.07\textwidth}}
\hline
\textbf{Hyperparameter} & \textbf{Value} \\ \hline
LSTM-State Size         & 100            \\ \hline
LSTM Layers             & 2              \\ \hline
Dropout                 & 0.68           \\ \hline
Second Dropout          & 0.5            \\ \hline
Epochs                  & 10             \\ \hline
Optimizer               & Adam           \\ \hline
CNN Kernel Size         & 3              \\ \hline
CNN Output Size         & 64             \\ \hline
\end{tabular}
}
\caption{Hyperparameter settings for trainingELMo-BiLSTM-CNN.}
\label{tab:BiLSTM-hyperparameter-settings}
\end{table}

\subsubsection{UniversalNER} 
The second method we explored is the fine-tuned UniversalNER approach, which leverages instruction tuning for named entity recognition (NER) tasks. UniversalNER is a versatile model, pre-trained for broad application scenarios such as open-domain NER. In our experiments, we initially employed a zero-shot approach, where the model was evaluated on our dataset without prior exposure. Following the zero-shot evaluation, we proceeded to fine-tune the model to optimize its performance.
We adapted a different methodology for the fine-tuning process due to the limitations encountered with the original approach proposed by Zhou et al. \citep{ZhouEtAl2024}. Their approach suggested retraining a LLaMA model on a UniversalNER-type dataset, but this was not feasible due to out-of-memory (OOM) constraints. Consequently, we introduced an alternative fine-tuning strategy better suited to our computational resources.
To prepare our dataset for this new fine-tuning approach, we first modified the data following the principles of conversation-style instruction tuning (original template), which is foundational to the UniversalNER model. The dataset was reformatted to align with the structure (template) expected by the UniversalNER architecture and its underlying LLaMA model. The template is illustrated in Figure~\ref{fig:UniversalNER}.  To convert our dataset to the Conversation-style template, we assigned the normalized text to the \(X_{passage}\) variable, \(t_T\) is the entity name (for example, product), and \(y_T\) is the entity for that product page (for example, 10g High Quality Hash). We generated the conversation-style template in JSON format based on these mappings. Once the dataset was appropriately formatted, we conducted a train-test split, allocating 80\% of the data for training and 20\% for testing.

For the new fine-tuning approach tailored to the UniversalNER model, we first started by analyzing the underlying structure of the UniversalNER model. The UniversalNER architecture describes a variant of the LLaMA model introduced by Touvron et al. \citep{TouvronEtAl2023}. Below, we will introduce and explain all the different components of the proposed UniversalNer model:

\begin{itemize}
  \item \textbf{Embedding Layer:} The model utilizes an embedding layer of 32,000 for the vocabulary and an embedding dimension of 4096. The padding id is set to zero, which indicates that index- in the vocabulary is used for padding. The large embedding size enables the model to capture various tokens and contextual meanings.
  \item \textbf{LlamaDecoderLayer:} The model core consists of 32 identical LlamaDecoderLayers. This characteristic implies a deep network capable of handling complex dependencies in text. 
  \item \textbf{LlamaSdpaAttention:} This layer is the scaled dot-product attention mechanism optimized for efficiency and performance. Each component (\(q_{proj}\), \(k_{proj}\), \(v_{proj}\), \(o_{proj}\)) is a linear layer with an input and output feature size of 4096, which matches the embedding dimension. The absence of bias in these layers can lead to more stable training as it reduces the number of parameters and potential overfitting. In the fine-tuning process, these linear layers will be targeted.
  \item \textbf{LlamaRotaryEmbedding:} This component is used within the attention mechanism to provide relative positional information, enhancing the ability of the model to understand the sequence and context without explicitly encoding absolute positions.
  \item \textbf{LlamaMLP:} This is a feedforward neural network within each decoder layer. It uses a gated linear unit structure with a Sigmoid Linear Unit as the activation function. The intermediate dimension is 11008, significantly larger than the input/output dimension (4096). This expansion and subsequent projection back help the model to capture non-linear relationships effectively.
  \item \textbf{LlamaRMSNorm:} RMS Normalization is applied before and after the attention mechanism. This helps stabilize the layer inputs and outputs, leading to more stable and faster training.
  \item \textbf{Output layer (lm head):} This linear layer maps the 4096-dimensional output of the last decoder back to the vocabulary size (32000). The absence of bias helps reduce the number of trainable parameters and might contribute to a more generalized model.
\end{itemize}

Understanding the architecture of the UniversalNER model allows us to tailor the fine-tuning process effectively. According to Zhou et al., \citep{ZhouEtAl2024}, training the original UniversalNER model requires substantial computational resources (e.g., 8 x A100 80GB GPUs). However, due to resource constraints and out-of-memory (OOM) issues, we developed a more resource-efficient fine-tuning strategy.
Our fine-tuning methodology includes fine-tuning UniversalNER via the Parameter-Efficient Fine-tuning (PEFT) method Low-Rank Adaptation (LoRA) introduced by Hu et al.\citep{HuEtAl2021}. LoRA reduces the computational burden by keeping most of the model’s parameters frozen and only updating two smaller low-rank matrices. This technique significantly decreases memory requirements while maintaining model performance. LoRA utilizes low-rank decomposition to update only two smaller matrices instead of the entire weight matrix.  Furthermore, LoRA is highly effective for adapting pre-trained models to new tasks without sacrificing their generalization capabilities. To further mitigate OOM issues, we applied k-bit quantization, which reduces the memory footprint of model weights during training.

To adapt our dataset for the fine-tuning process, we adjusted the chat-style template of the tokenizer associated with the UniversalNER model to match the structure of our dataset. This ensures that the tokenizer correctly interprets the keys of our chat data. This new template is applied across all chat conversations in the training and test dataset. The new template uniformly processes all entries, preparing them for the following tokenization step. Finally, we map a tokenization function across the train and test datasets. This step transforms the textual data into numerical tokens necessary for fine-tuning with LoRA. 

For the supervised fine-tuning of the UniversalNER model, we used the SFTTrainer developed by HuggingFace. For the fine-tuning process, the parameters mentioned in Table \ref{tab:SFTTrainer-parameters} are used as these performed the best while not causing any OOM issues. After fine-tuning the model, we merge the new model with the original UniversalNER model. Merging combines the broad understanding of NER from the original model with the specialized insights from the fine-tuned model. This helps prevent the fine-tuned model from overfitting to our data while retaining the expansive knowledge of the original. In essence, this process allows for a transfer of updated weights from the fine-tuned model, selectively updating parts of the original model that have been refined during fine-tuning.

\begin{table}[h!]
\fontsize{9pt}{9pt}\selectfont
\centering
{\def\arraystretch{1.5}
\scalebox{0.77}{
\begin{tabular}{p{0.4\textwidth}p{0.16\textwidth}}
\hline
\textbf{Parameter} & \textbf{Value} \\ \hline
\multicolumn{2}{|c|}{\textbf{LoRA Parameters}}\\ \hline
LoRA attention dimension (\texttt{lora\_r}) & 64 \\ \hline
Alpha parameter for LoRA scaling (\texttt{lora\_alpha}) & 16 \\ \hline
\multicolumn{2}{|c|}{\textbf{bitsandbytes Parameters}} \\ \hline
Activate 4-bit precision base model loading (\texttt{use\_4bit}) & True\\ \hline
\multicolumn{2}{|c|}{\textbf{TrainingArguments Parameters}} \\ \hline
Number of training epochs (\texttt{num\_train\_epochs}) & 1 \\ \hline
Enable fp16/bf16 training (\texttt{bf16}) & False \\ \hline
Batch size per GPU for training (\texttt{per\_device\_train\_batch\_size}) & 2 \\ \hline
Enable gradient checkpointing (\texttt{gradient\_checkpointing}) & True \\ \hline
Initial learning rate (\texttt{learning\_rate}) & 2e-4 \\ \hline
Optimizer (\texttt{optim}) & paged\_adamw\_32bit \\ \hline
Learning rate schedule (\texttt{lr\_scheduler\_type}) & constant \\ \hline
\multicolumn{2}{|c|}{\textbf{SFT Parameters}} \\ \hline
Maximum sequence length to use (\texttt{max\_seq\_length}) & 40 \\ \hline
Pack multiple short examples in the same input sequence\texttt{packing}) & True \\ \hline
\end{tabular}
}
}
\caption{Key Parameter Settings for Fine-tuning UniversalNER.}
\label{tab:SFTTrainer-parameters}
\end{table}

\subsubsection{GLiNER} 
\begin{figure}[h!]
    \centering
    \includegraphics[width=0.4\textwidth]{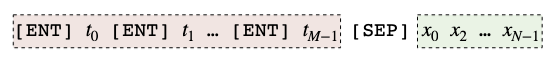}
    \caption{Input format GLiNER model \citep{ZaratianaEtAl2023}.}
    \label{fig:GLiNER-input}
\end{figure}

For the final approach, we adopted the GLiNER model, as introduced by Zaratiana et al. \citep{ZaratianaEtAl2023}. To fine-tune the GLiNER model, we first adapted our dataset to conform to the model’s input requirements. The input format consists of a unified sequence that merges natural language descriptions of entity types with the text from which the entities are extracted. The input format is presented in Figure \ref{fig:GLiNER-input}. 

In Figure \ref{fig:GLiNER-input}, the [ENT] token serves as a marker preceding each entity type, while the [SEP] token acts as a delimiter, dividing the entity type sequence from the input text. The unified input is processed by the token encoder to compute interactions among all tokens (both entity types and input text), generating contextualized representations. The encoder's output value for each entity type, derived from all the [ENT] token representations, is determined using the following formula: \(p = \{p_i\}_{i=0}^{M-1} \in \mathbb{R}^{M \times D}\). In the same way, the representations of each word in the input text are calculated by: \(h = \{h_i\}_{i=0}^{N-1} \in \mathbb{R}^{N \times D}\). 
For words tokenized into subwords, we adopt the standard practice in NER literature by using the representation of the first subword.

The GLiNER model aims to create a unified latent space by encoding entity types and span embeddings. The representation of an entity is calculated utilizing a two-layer feedforward network (FFN) to refine the original representation \(p\); this results in the following equation \(q = \{q_i\}_{i=0}^{M-1} \in \mathbb{R}^{M \times D}\) \citep{ZaratianaEtAl2023}. In the GLiNER model,  \(q_{ij} \in \mathbb{R}^{D}\) denotes a span that starts at position i and finishes at position j. This span is utilized in the input text and is calculated as \(S_{ij} = \text{FFN}(h_i \otimes h_j)\). In this equation, a two-layered feedforward network is denoted as FFN, and the concatenation process is stated as \(\otimes\). The calculations presented above can be executed in parallel when circumstances permit. In addition, we set the order to keep linear complexity without harming Recall. 

For analyzing if a span \((i,j)\) matches a specific entity with type t, the following score is calculated: \(\phi(i,j,t) = \sigma(S_{ij}^T q_t) \in \mathbb{R}
\). In this formula, \(\sigma\) represents a Sigmoid activation function, and the probability of a span belonging to type \(t\) is expressed as \(\phi(i,j,t) \).

During training and fine-tuning, our objective is to maximize the number of correct span-type matches (positive pairs) while minimizing incorrect matches (negative pairs) by optimizing the model's hyperparameters. Following Zaratiana et al. \citep{ZaratianaEtAl2023}, a span \((i, j)\) paired with an entity type \(t\) forms a positive pair  \((s \in \mathbb{P})\) if the span is labeled with type \(t\) in the training data. Otherwise, it is considered a negative pair \((s \in \mathbb{N})\). The training loss for a single example, consisting of spans \(\mathbb{S}\) and entity types \(\mathbb{T}\) is calculated as \(\mathcal{L}_{\text{BCE}} = - \sum_{s \in S \times T} I_{s \in P} \log(\phi(s)) + I_{s \in N} \log(1 - \phi(s))''\). 

In the training loss function, \(s\) denotes a pair (positive or negative) of span/entity types, and \(\mathbb{I}\) represents an indicator function, meaning it returns 1 when the specified condition is true and 0 otherwise \citep{ZaratianaEtAl2023}. The loss function of GLiNER is complementary to the binary cross entropy.


Fine-tuning the GLiNER model involves a meticulous process leveraging PyTorch's flexible framework, combined with additional tools such as accelerate and transformers, to ensure efficient training across different hardware configurations. The hyperparameters used in our fine-tuning experiments are detailed in Table \ref{tab:GLiNER-hyperparameters}. This fine-tuning setup was designed to balance performance improvements with computational efficiency, addressing the challenges posed by resource limitations.

\begin{table}[h!]
\fontsize{9pt}{9pt}\selectfont
\centering
{\def\arraystretch{1.5}
\scalebox{0.8}{
\begin{tabular}{p{0.45\textwidth}p{0.05\textwidth}}
\hline
\textbf{Parameter} & \textbf{Value} \\
\hline
\multicolumn{2}{|c|}{\textbf{General Training Parameters}} \\
\hline
Number of steps (\texttt{num\_steps}) & 500 \\
\hline
Training batch size (\texttt{train\_batch\_size}) & 2 \\
\hline
Evaluation frequency (\texttt{eval\_every}) & 10 \\
\hline
Directory for saving logs (\texttt{save\_directory}) & ``logs'' \\
\hline
Warmup ratio (\texttt{warmup\_ratio}) & 0.1 \\
\hline
Device for training (\texttt{device}) & `cpu' \\
\hline
Learning rate for encoder (\texttt{lr\_encoder}) & 1e-5 \\
\hline
Learning rate for other parameters (\texttt{lr\_others}) & 5e-5 \\
\hline
Freeze token representation (\texttt{freeze\_token\_rep}) & False \\
\hline
\multicolumn{2}{|c|}{\textbf{Data Augmentation and Processing}} \\
\hline
Maximum types (\texttt{max\_types}) & 25 \\
\hline
Shuffle types (\texttt{shuffle\_types}) & True \\
\hline
Random drop of types (\texttt{random\_drop}) & True \\
\hline
Maximum negative type ratio (\texttt{max\_neg\_type\_ratio}) & 1 \\
\hline
Maximum sequence length (\texttt{max\_len}) & 3000 \\
\hline
\end{tabular}
}
}
\caption{Key parameter settings for fine-tuning GLiNER.}
\label{tab:GLiNER-hyperparameters}
\end{table}

\section{Results}
While doing NER, both the boundaries and types of entities are determined. As stated by Shah et al. \citep{ShahEtAl2022}, a named entity is correctly identified in ``exact-match evaluation'' when its boundaries and type are consistent with the ground truth. Since we are only interested in the information linked to the product on the product listing page, this becomes even more relevant. 
The evaluation metrics used include Precision, Recall, and the F-1 Score.

\subsection{ELMo-BiLSTM Evaluation}
\begin{figure}[h!]
    \centering
    \includegraphics[width=0.95\linewidth]{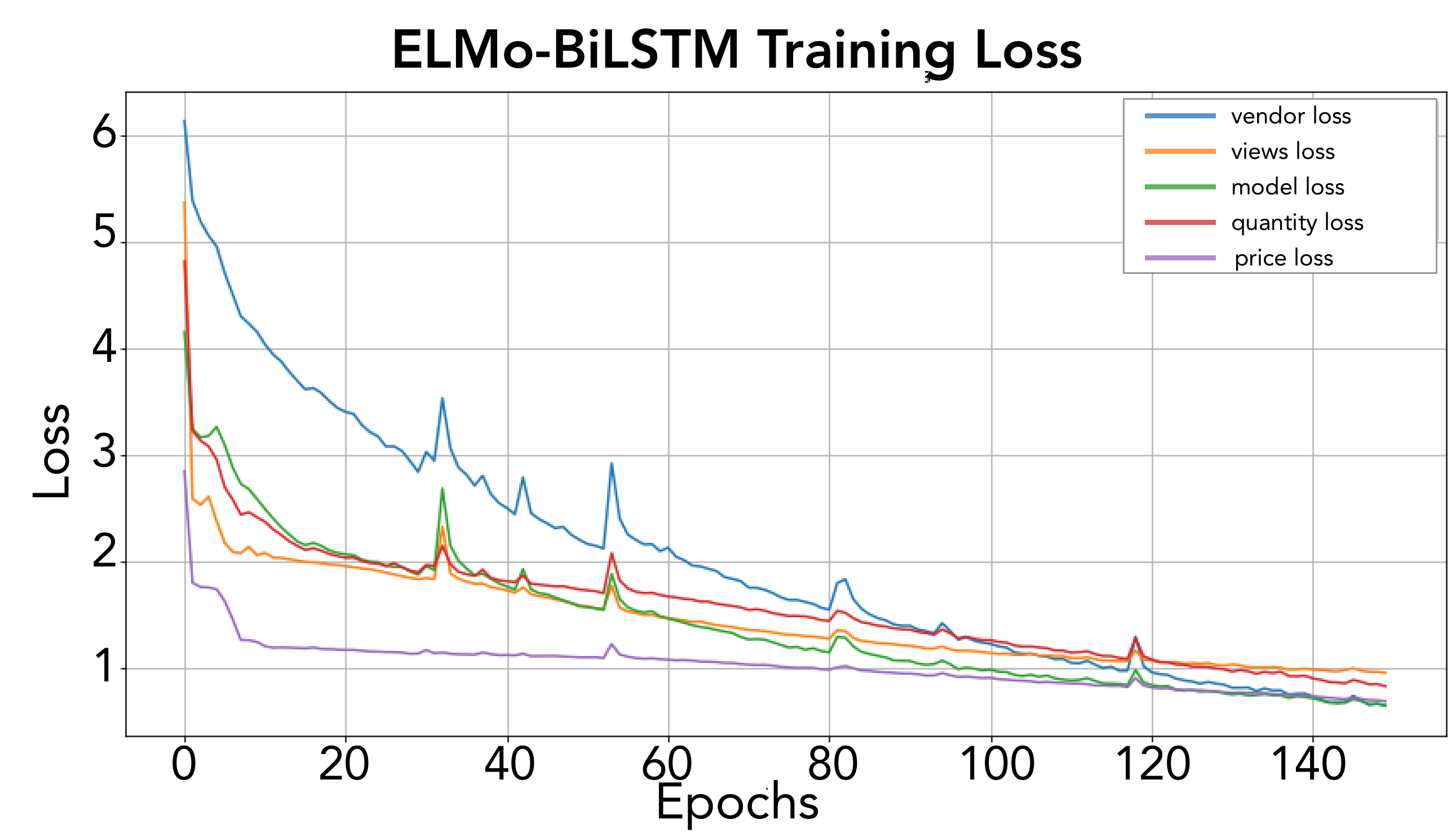}
  \captionof{figure}{Loss per epoch.}
  \label{fig:Bi-LSTM-trainingA}
\end{figure}

\begin{figure}[h!]
    \centering
      \includegraphics[width=0.95\linewidth]{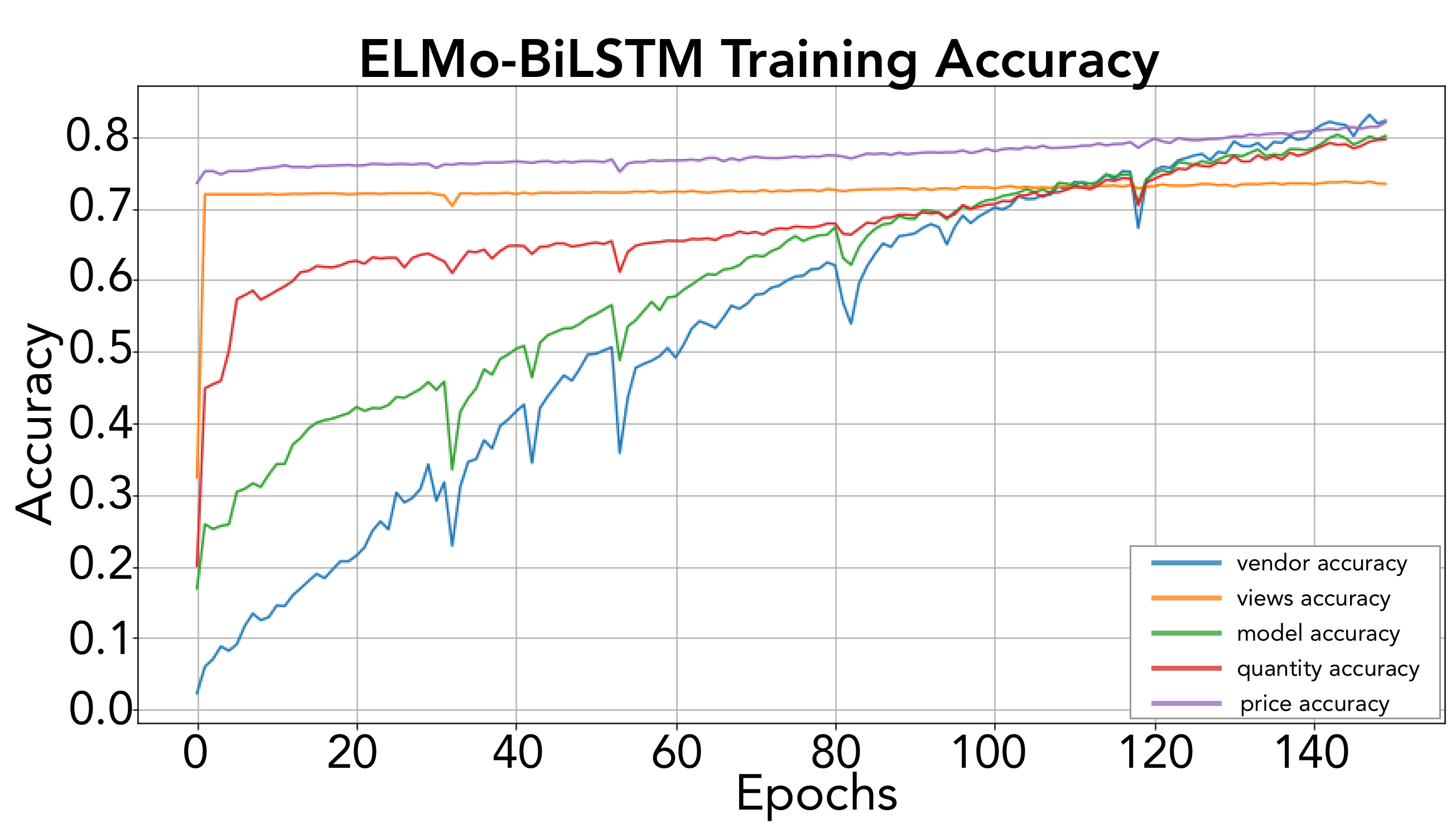}
  \captionof{figure}{Accuracy per epoch.}
  \label{fig:Bi-LSTM-trainingD}
\end{figure}

\begin{figure}[h!]
    \centering
     \includegraphics[width=0.95\linewidth]{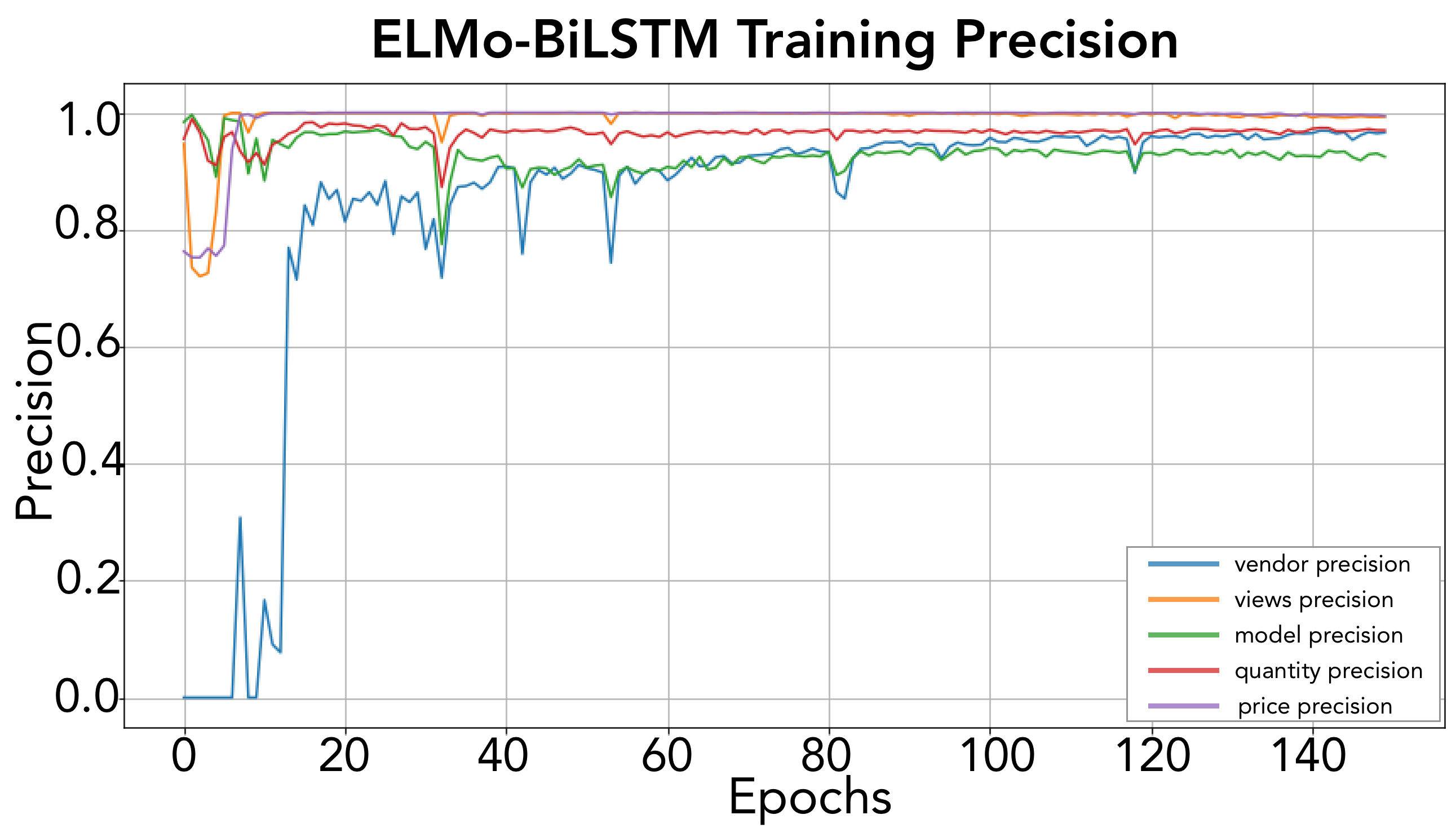}
  \captionof{figure}{Precision per epoch.}
  \label{fig:Bi-LSTM-trainingC}
\end{figure}

\begin{figure}[h!]
    \centering
    \includegraphics[width=0.95\linewidth]{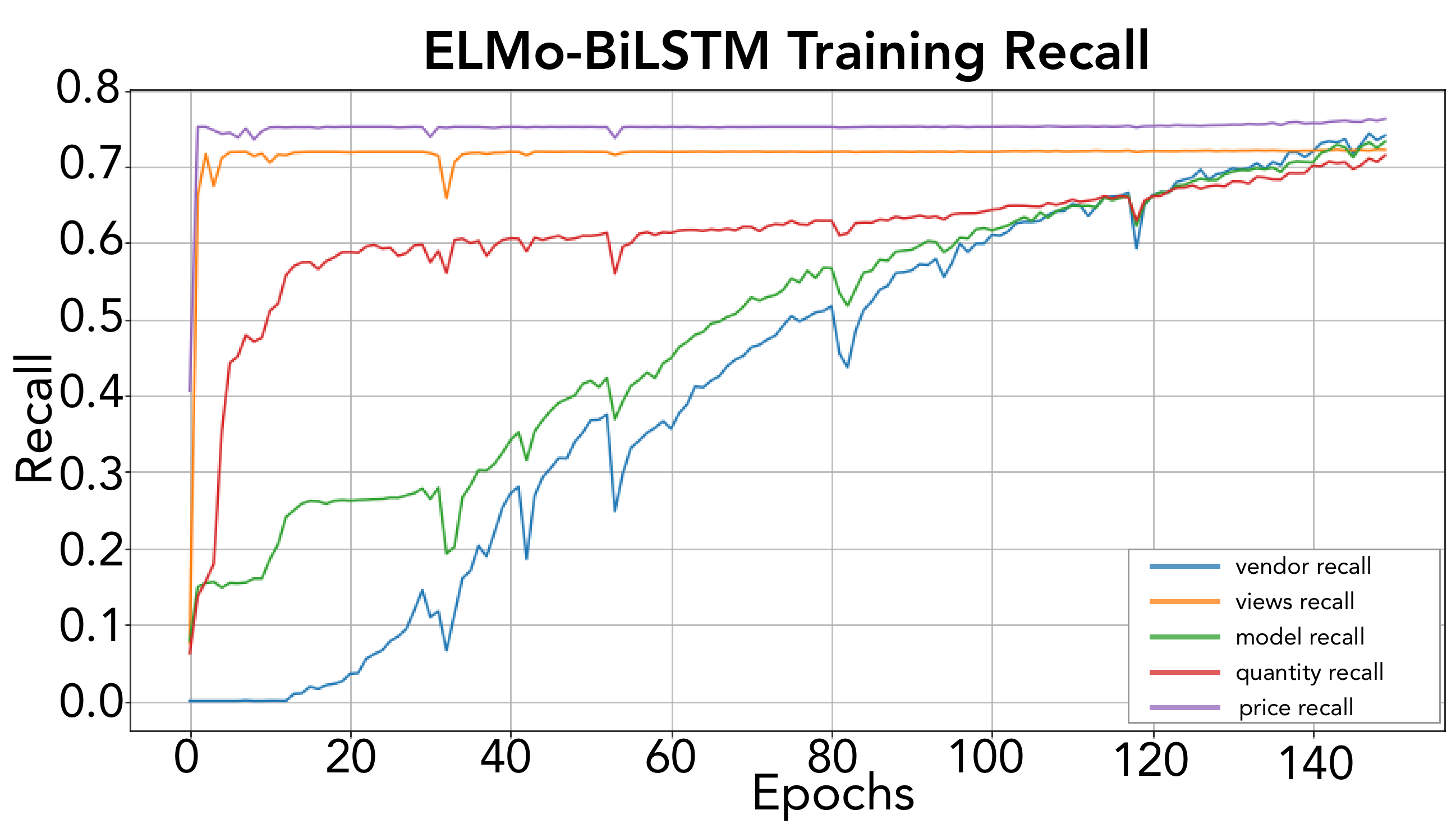}
  \captionof{figure}{Recall per epoch.}
  \label{fig:Bi-LSTM-trainingB}
\end{figure}

\begin{figure}[h!]
    \centering
     \includegraphics[width=0.95\linewidth]{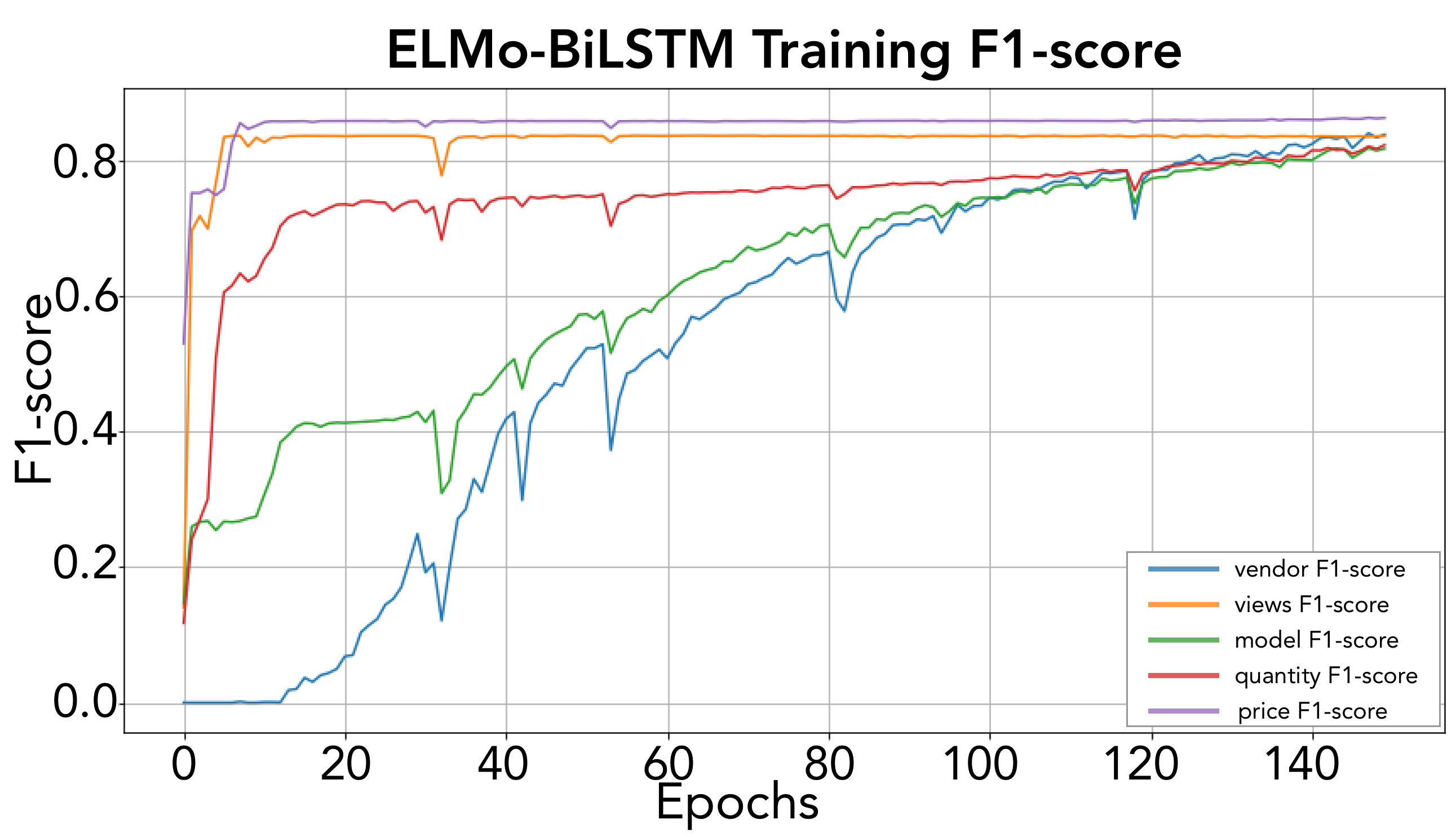}
  \captionof{figure}{F1-score per epoch.}
  \label{fig:Bi-LSTM-trainingE}
\end{figure}

After preparing the data, we developed and trained the ELMo-embedded BiLSTM model designed to extract the complex entities identified in product listing pages from DNM. We trained and developed the BiLSTM-CNN model with ELMo embedding using the hyperparameters from Section \ref{experiments}. The plots in Figure~\labelcref{fig:Bi-LSTM-trainingA,fig:Bi-LSTM-trainingB,fig:Bi-LSTM-trainingC,fig:Bi-LSTM-trainingD,fig:Bi-LSTM-trainingE} depict the training dynamics and the model's performance across various entity categories. The y-axis in these figures represents the performance metric, while the x-axis indicates the number of epochs. Each plot illustrates the model's progress in predicting specific entity types.


As shown in Figure \ref{fig:Bi-LSTM-trainingA}, the training loss for all outputs consistently decreases over the epochs, indicating that the model effectively learns from the training data. Notably, entities such as ``vendor'', ``views'', and ``quantity in stock'' exhibit higher initial losses, suggesting that these features were more challenging for the model to predict at the outset. However, the substantial reduction in loss over time demonstrates that the model successfully adapts to these complexities. In contrast, the entities ``category'', ``model'', and ``price'' show more gradual and stable decreases in loss, suggesting smoother learning curves for these features.

The accuracy trends in Figure \ref{fig:Bi-LSTM-trainingD} show the observations from the loss curves. The ``views'' and ``quantity in stock'' entities display steady and significant improvements in accuracy, ultimately achieving the highest scores. This suggests that the model becomes increasingly adept at recognizing patterns related to these entities. The ``vendor'' entity, which begins with a lower accuracy, also shows consistent improvement, highlighting the model's ability to learn more challenging predictions through sustained training. Similar trends are observed for the ``model'' entity.

Figure \ref{fig:Bi-LSTM-trainingC} presents the precision metrics, which initially fluctuate before stabilizing across most entity types. The ``price'' entity consistently maintains high precision throughout the training process, indicating that the model accurately predicts positive instances of this feature. This stability suggests that price-related data may have more distinguishable patterns, making it easier for the model to learn. On the other hand, the ``vendor'' entity shows greater variability in precision, likely due to the vendor data's complexity or irregularity, making precise predictions more challenging.

As shown in Figure \ref{fig:Bi-LSTM-trainingB}, the recall scores steadily improve across all entities, reflecting the model's growing ability to identify relevant instances. The higher recall for the ``views'' and ``price'' entities suggests that the model is particularly effective at retrieving these features, likely due to more consistent labeling or inherent characteristics that make these entities more distinguishable in the data.

The F1 scores, which balance precision and recall, reinforce the abovementioned observations. Figure \ref{fig:Bi-LSTM-trainingE} shows an upward trend in F1 scores across all entities, with ``price'' and ``views"'' emerging as the leading entities during the later stages of training. The substantial improvement in the F1 score for the ``vendor'' entity underscores the model's increasing capability to accurately predict this complex feature over time.

\begin{figure}[hbt!]
    \centering
    \includegraphics[width=0.45\textwidth]{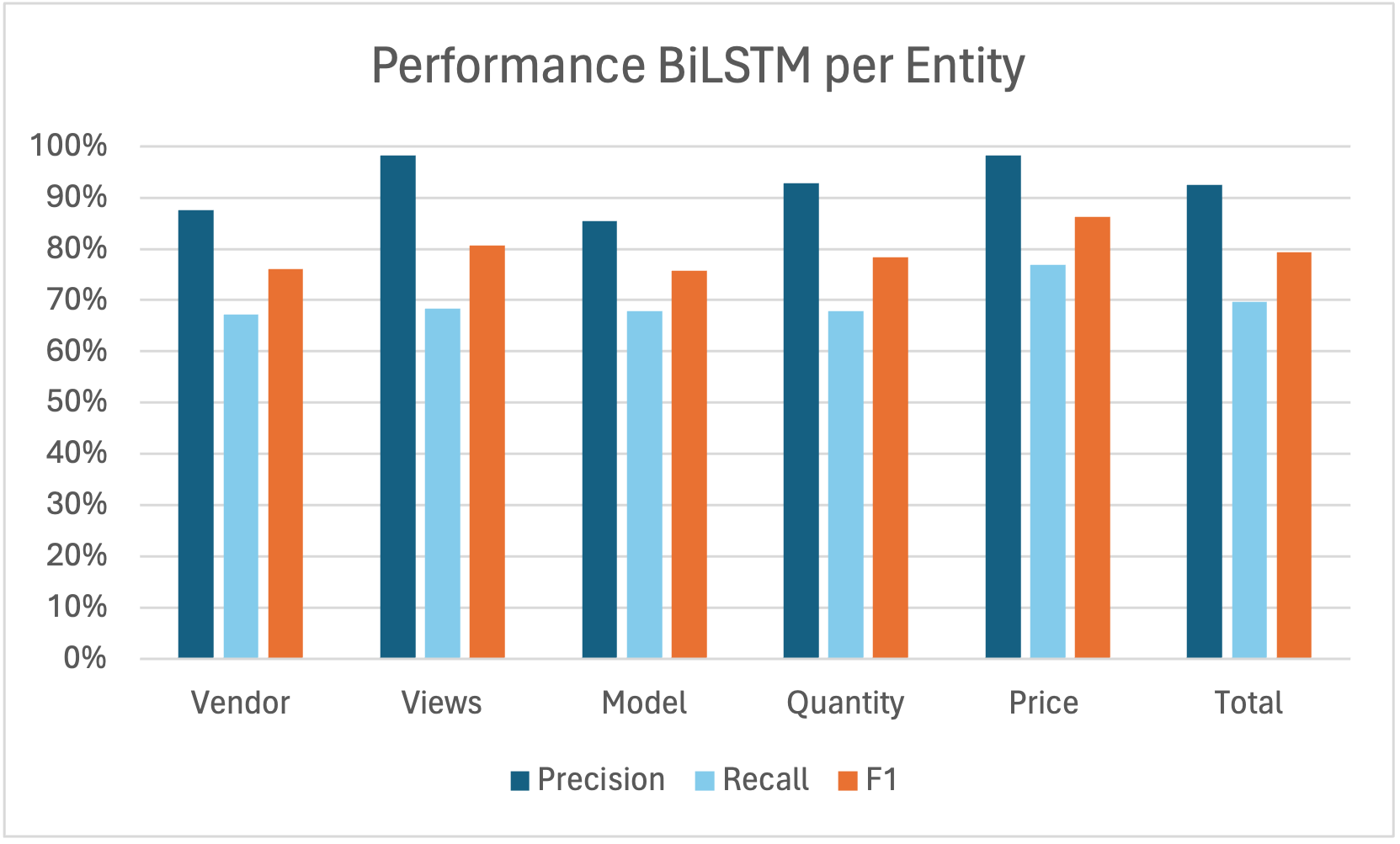}
    \caption{ELMo-BiLSTM key performance indicators per entity.}
    \label{fig:BiLSTM-performance}
\end{figure}

Upon completing the training, we evaluated the model's performance on the test dataset, as shown in Figure \ref{fig:BiLSTM-performance}. The results demonstrate the model's efficacy in extracting key entities from DNM product listings, with varying performance levels across different entity types.

\subsection{Zero-shot Approach Evaluation}
\begin{figure}[hbt!]
    \centering
    \includegraphics[width=0.5\textwidth]{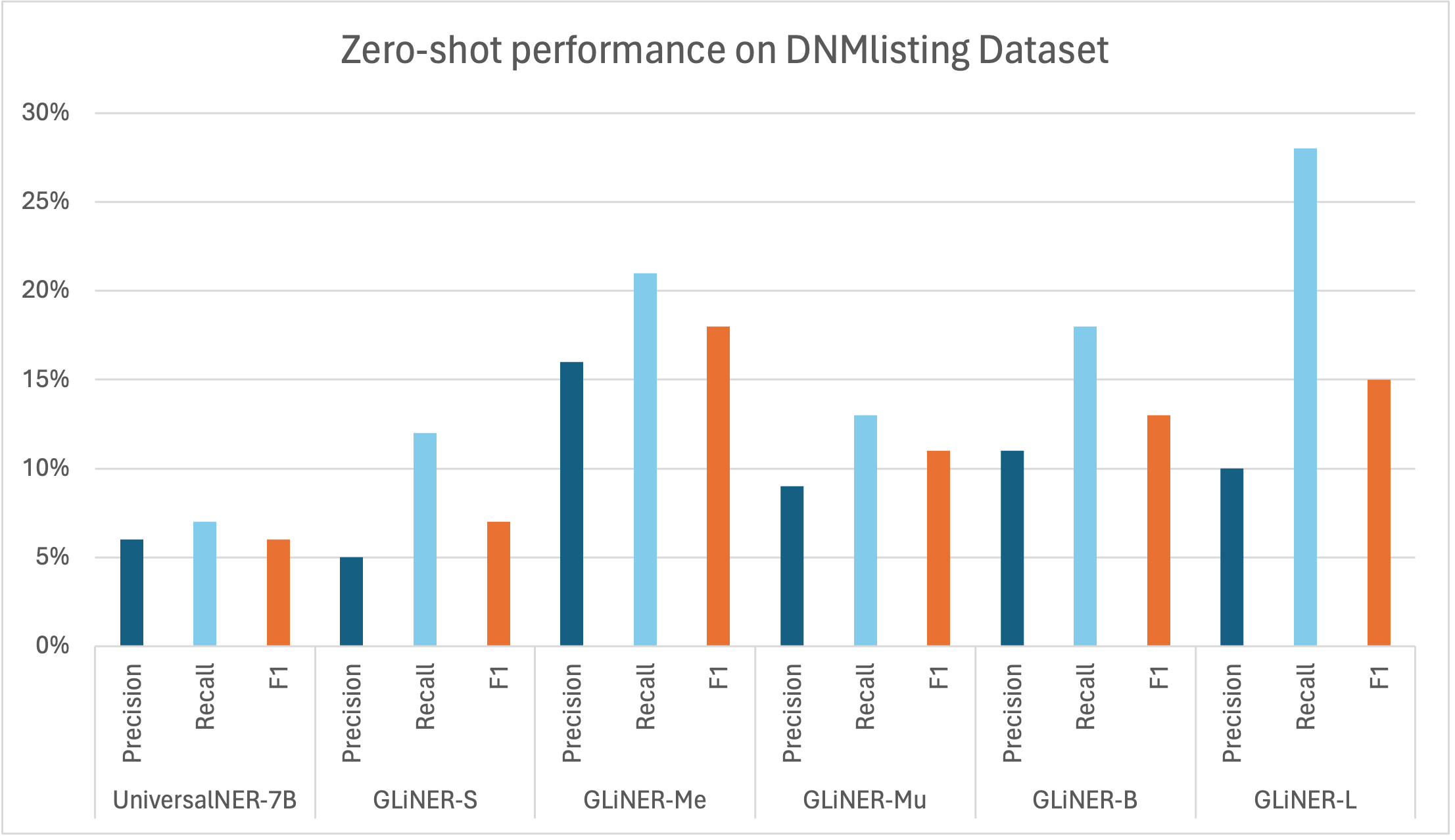}
    \caption{Zero-shot key performance indicators per model on the DNMlistings Dataset.}
    \label{fig:zero-shot-performance}
\end{figure}

We now discuss the performance of the UniversalNER and GLiNER on our created dataset using their zero-shot approach. For this experiment, we utilized the pre-trained UniversalNER and GLiNER models without any task-specific fine-tuning, assessing their performance in a zero-shot setting. To provide context, we compared the results on our DNMsListing dataset with 20 other established NER datasets across various domains, including biomedical texts, news articles, and social media data (Table \ref{tab:performance_metrics_zeroshot}).
By adding the DNMsListing dataset to this benchmark, we were able to assess the relative performance of both models in an unstructured, challenging domain.
The values in Table \ref{tab:performance_metrics_zeroshot} show that GLiNER-L significantly outperforms UniversalNER-7B on our dataset. The score difference suggests that GLiNER-L is better at identifying and classifying the labeled entities in our dataset, as it has more than double the score of UniversalNER-7B.
It is important to note that the overall zero-shot performance of both models on the DNMsListing dataset was lower compared to their performance on other NER datasets. This decline in performance can likely be attributed to the unique characteristics of Darkweb content. The text in DNM product listings often exceeds the token length typically encountered in surface web datasets, leading to truncation and, consequently, a reduction in model effectiveness. Additionally, Darkweb data presents further challenges, such as increased variability in spelling, sentence structure, and formatting, which are not prevalent in more standardized datasets. These findings underscore the need for specialized models capable of handling the idiosyncrasies of Darkweb content.
A more detailed analysis of zero-shot performance on the DNMsListing dataset is presented in Figure \ref{fig:zero-shot-performance}, where several key insights emerge regarding the comparative effectiveness of the models. GLiNER-L consistently outperforms the other models across all evaluation metrics, achieving superior Precision and F1 scores. This performance suggests that GLiNER-L exhibits strong generalization capabilities in zero-shot scenarios, efficiently identifying relevant entities while minimizing both false positives and false negatives.
The performance of GLiNER-Me and GLiNER-Mu, while commendable, did not reach the levels exhibited by GLiNER-L. GLiNER-Me demonstrated better precision and F1 scores compared to GLiNER-Mu, though GLiNER-Mu showed a slight edge in recall, indicating its strength in capturing a broader range of relevant entities. GLiNER-S, although outperforming UniversalNER-7B, lagged behind the other GLiNER variants, suggesting that it may require further optimization for zero-shot learning tasks.
On the other hand, UniversalNER-7B demonstrated significantly lower performance compared to all GLiNER variants. Its relatively poor results on the DNMsListing dataset suggest potential limitations in its architecture or pre-training methodologies when applied to the unique challenges presented by Darkweb content in zero-shot settings. These results suggest that further refinement of UniversalNER's design may be necessary to enhance its performance in more complex and unstructured domains like the Darkweb.
Overall, these findings highlight the superior performance of the GLiNER models, particularly GLiNER-L, in zero-shot entity recognition on challenging datasets. They also emphasize the necessity for targeted model adjustments to accommodate the diverse nature of Darkweb data, where conventional zero-shot NER models may struggle.


\begin{table}[h!]
\centering
\begin{tabular}{lccc}
\toprule
\textbf{Dataset} & \textbf{UniversalNER-7B} & \textbf{GLiNER-L} \\
\midrule
ACE05           & 36.9 & 27.3 \\
AnatEM          & 25.1 & 33.3 \\
bc2gm           & 46.2 & 47.9 \\
bc4chemd        & 47.9 & 43.1 \\
bc5cdr          & 68.0 & 66.4 \\
Broad Tweeter   & 67.9 & 61.2 \\
CoNLL03         & 72.2 & 64.6 \\
FabNER          & 24.8 & 23.6 \\
FindVehicle     & 22.2 & 41.9 \\
GENIA           & 54.1 & 55.5 \\
HarveyNER       & 18.2 & 22.7 \\
MIT Movie       & 42.4 & 57.2 \\
MIT Restaurant  & 31.7 & 42.9 \\
MultiNERD       & 59.3 & 59.7 \\
ncbi            & 60.4 & 61.9 \\
OntoNotes       & 27.8 & 32.2 \\
PolyglotNER     & 41.8 & 42.9 \\
TweetNER7       & 42.7 & 41.4 \\
WikiANN         & 55.4 & 58.9 \\
WikiNeural      & 69.2 & 71.8 \\
DNMsListing      & 6.3 & 15.0 \\
\midrule
\textbf{Average} & 45.7 & 47.8 \\
\bottomrule
\end{tabular}
\caption{Zero-shot performance on 20 NER datasets.
Results of UniversalNER are reported by Zhou et al. \citep{ZhouEtAl2024}, and GLiNER is reported by Zaratiana et al. \citep{ZaratianaEtAl2023}.}
\label{tab:performance_metrics_zeroshot}
\end{table}

\subsection{In-domain Fine-tuning}
\begin{figure}[hbt!]
    \centering
    \includegraphics[width=0.45\textwidth]{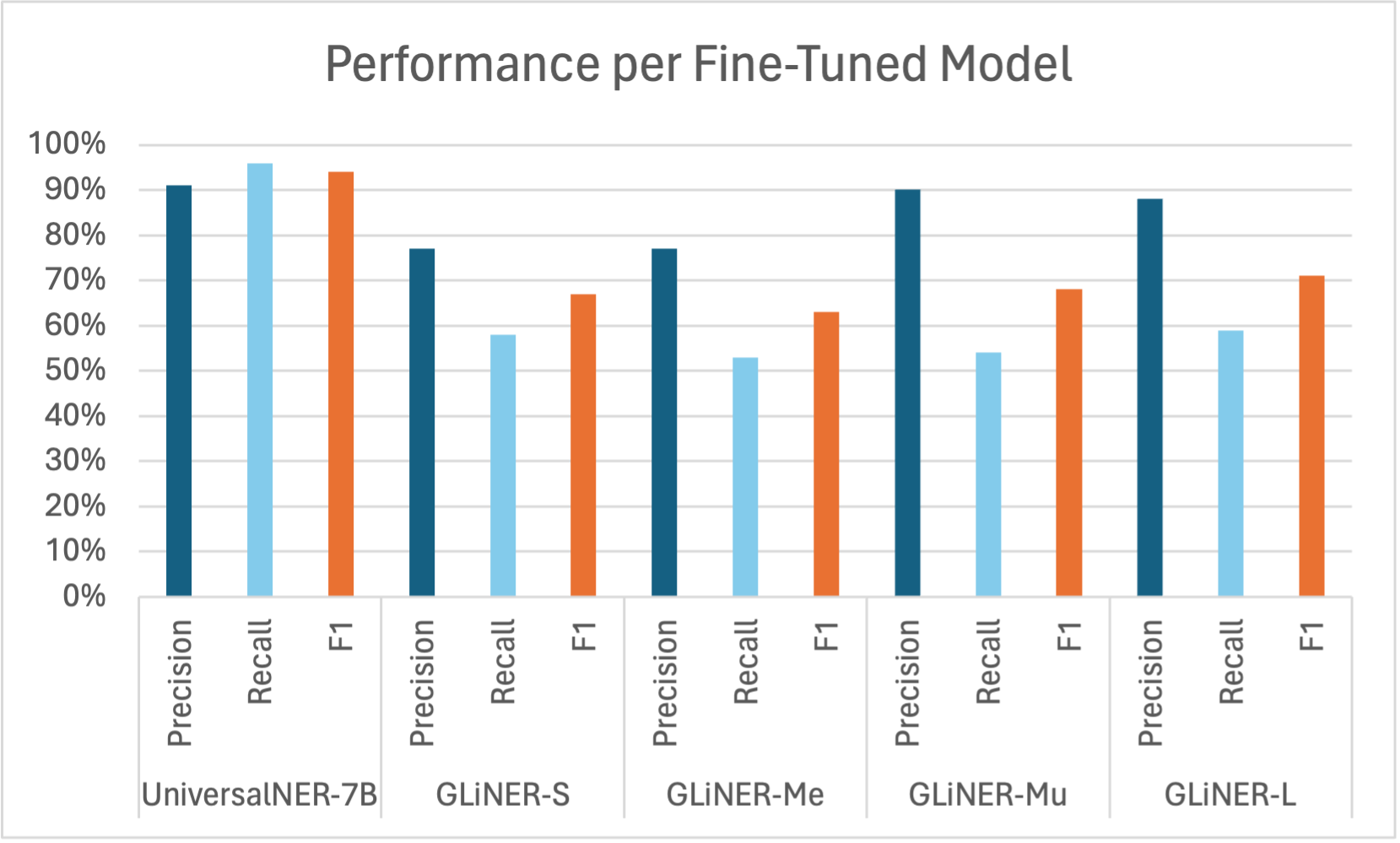}
    \caption{Key performance indicators per model after fine-tuning.}
    \label{fig:finetuned-performance}
\end{figure}

Both zero-shot NER models have the ability to improve entity extraction in domain-specific contexts by finetuning. Our evaluation of the fine-tuning process aligns with the claims made by Zaratiana et al. \citep{ZaratianaEtAl2023}, who noted that UniversalNER requires significantly more time and computational resources than GLiNER. However, the gap between the two models has narrowed with our new fine-tuning proposed approach. We successfully finetuned the UniversalNER model on 1 NVIDIA A100 40GB. Compared with the 8 NVIDIA A100 80GB needed that Zhou et al. \citep{ZhouEtAl2024} described, this is a substantial improvement. In addition, the training time was reduced to approximately 26 minutes. If we compare these resources to those needed for finetuning GLiNER, GLiNER is much more efficient. We finetuned GLiNER on a CPU instead of specifically needing a GPU for finetuning as by UniversalNER. The time for finetuning GLiNER depends on the size of the model. Figure \ref{fig:finetuned-performance} presents the performance of the finetuned models on the test dataset. UniversalNER-7B achieves impressive results with a Precision of 91\%, Recall of 96\%, and an F1 score of 94\%, indicating strong Accuracy and Recall balance. In contrast, the GLiNER models exhibit more variable performance with generally lower scores. GLiNER Small, for instance, has a Precision of 77\% and Recall of 58\%, causing an F1 score of 67\%. This performance suggests reasonable accuracy but a limited ability to capture all relevant entities. GLiNER Medium and GLiNER Large have similar issues, with GLiNER Medium scoring slightly lower in recall and F1 score, indicating a challenge in identifying all relevant entities. While GLiNER Multi has a high Precision of 90\%, it suffers significantly in Recall at 54\%, resulting in an F1 score of 68\%, highlighting a critical gap in capturing a broader set of relevant entities. Finetuning the zero-shot NER tagging models enormously improves the ability to extract the correct entities. For the GLiNER approach, we conducted the finetuning on a smaller and less variable (one marketplace) as well; this led to a higher F1 score but was not in line with the goal of this research. 

\subsection{Proof of Concept Comparison}
\begin{figure}[hbt!]
    \centering
    \includegraphics[width=0.45\textwidth]{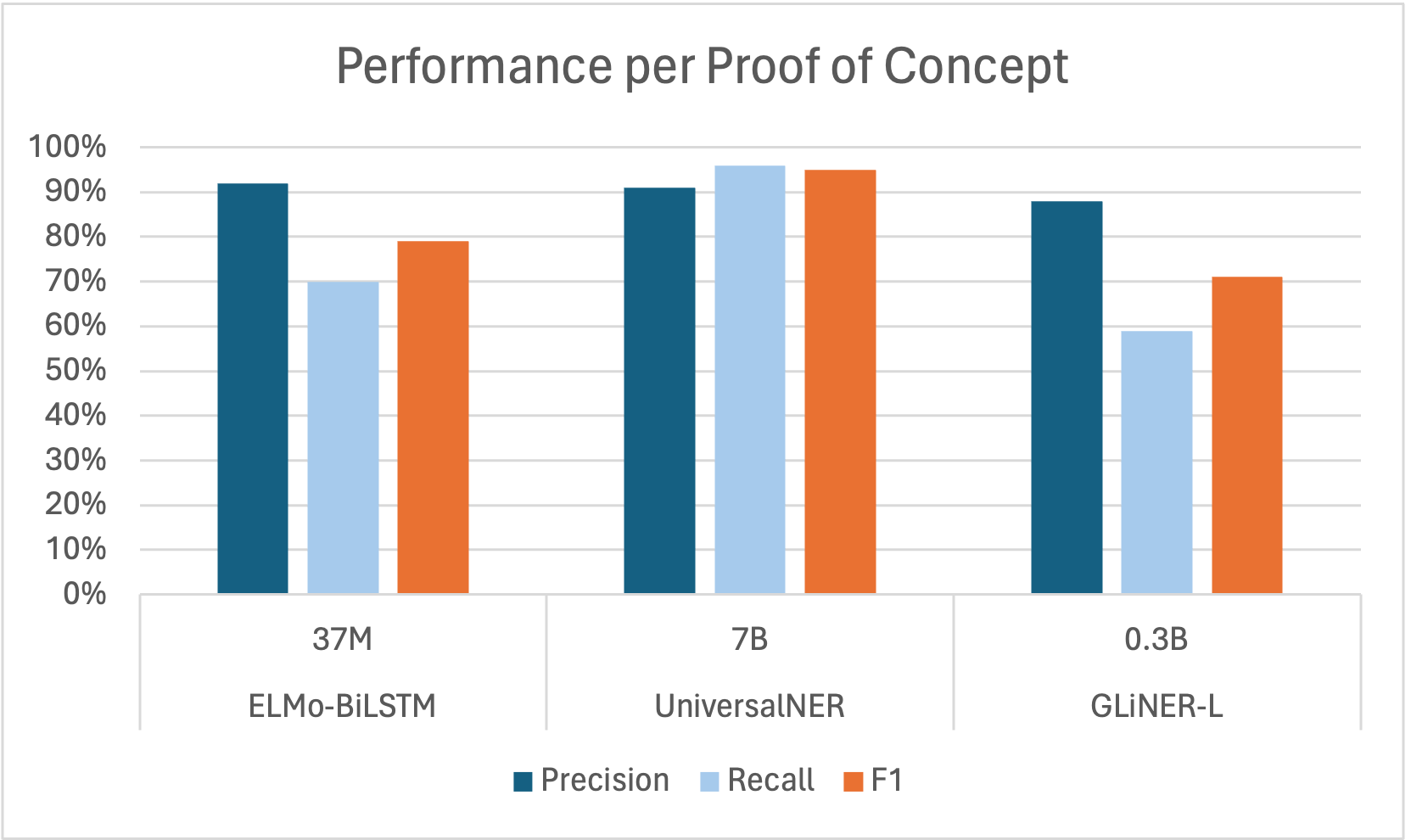}
    \caption{Key performance indicators per proof of Concept.}
    \label{fig:proof-of-concept-performance}
\end{figure}

In our study, we developed and evaluated three different proofs of concept specialized in retrieving entities from product listing pages from DNMs. In Figure \ref{fig:proof-of-concept-performance}, we evaluate the performance of three distinct proof of concepts: ELMo-BiLSTM, UniversalNER 7B, and GliNER-L. We developed the first proof of concept from scratch and finetuned the latter two on our DNMsListing dataset. Each concept varies significantly in the number of parameters, reflecting differences in architectural complexity and potential capabilities. The ELMo-BiLSTM model is the most miniature, containing approximately 37 million parameters, demonstrating high Precision at 92\%, suggesting it is highly effective at correctly identifying proper positive entities. Due to the lower Recall at 70\%, it indicates that the model misses a considerable number of relevant entities. The F1 score reflects this limitation by being 79\%, which, while respectable, suggests room for improvement in balancing Precision and recall. 

Finetuned UniversalNER-7B, the largest model with 7 billion parameters, shows a slight decrease in Precision (91\%) compared to the ELMo-BiLSTM approach but excels remarkably in the Recall at 96\%. This high recall rate indicates that UniversalNER-7B can identify nearly all relevant entities, substantially enhancing its F1 score to 95\%. The performance of this approach underscores the benefits of leveraging a more extensive parameter set to achieve depth and complexity in model learning, leading to superior overall efficacy in entity recognition tasks on content from the dark web. 

The finetuned GLiNER-L approach, with its 0.3 billion parameters, achieves a Precision of 88\%. However, its Recall is notably low at 59\%, the least effective among the models. This significant shortfall in Recall impacts its F1 score, which has a value of 71\%. We saw that specific entities, such as the ``Quantity in Stock'' and ``Product Views'', were complicated to retrieve from the zero-shot approach for the GLiNER model. Its architecture and fine-tuning process may limit its ability to generalize to new entities, affecting overall performance.

\subsection{Robustness comparison}
Robustness, in the context of dark web information extraction, is defined as the method's ability to maintain high performance across diverse web environments, withstand common challenges of the dark web, such as dynamic content, anti-scraping mechanisms, and variations in the data structure, and adapt to future web technologies and standards. We created an additional dataset to measure the robustness of the proof of concepts developed. The dataset for testing robustness consists of data from a different DNM, namely Palmetto State Armory Market. On the contrary of the other DNMs analyzed in this study, Palmetto sells products related to firearms instead of drugs. We used the same preprocessing methodology on the Palmetto dataset that we used to create the DNMsListings dataset. After the preprocessing, we utilized RegEx patterns to extract informative entities from the product listing web pages. As the primary goal of this dataset is to evaluate the robustness of the models developed, we added entries that were not in the dataset before, such as ``Stock Keeping Unit Number'' and ``Brand''. The dataset created covered 1386 product listings from the Palmetto market. The data included 196 unique vendors and 19 unique product categories, with an average of 1246 tokens per web page.

We did not evaluate the ELMo-embedded BiLSTM model and the GliNER model in terms of robustness due to the characteristics and performance of the model. Since the BiLSTM model does not have a zero-shot foundation, the model cannot predict new entities diverging from the data on which the model is trained. This leads to the BiLSTM model underperforming compared to the other approaches introduced. In addition, the embedding layer we proposed for the BiLSTM model cannot analyze and predict multi-token sequences. The GLiNER model is not included in the robustness evaluation as its performance after training was worse than the results of the UniversalNER approach. Because of this, we will analyze the performance of the fine-tuned UniversalNER approach in the context of robustness. 

\begin{figure}[hbt!]
    \centering
    \includegraphics[width=0.45\textwidth]{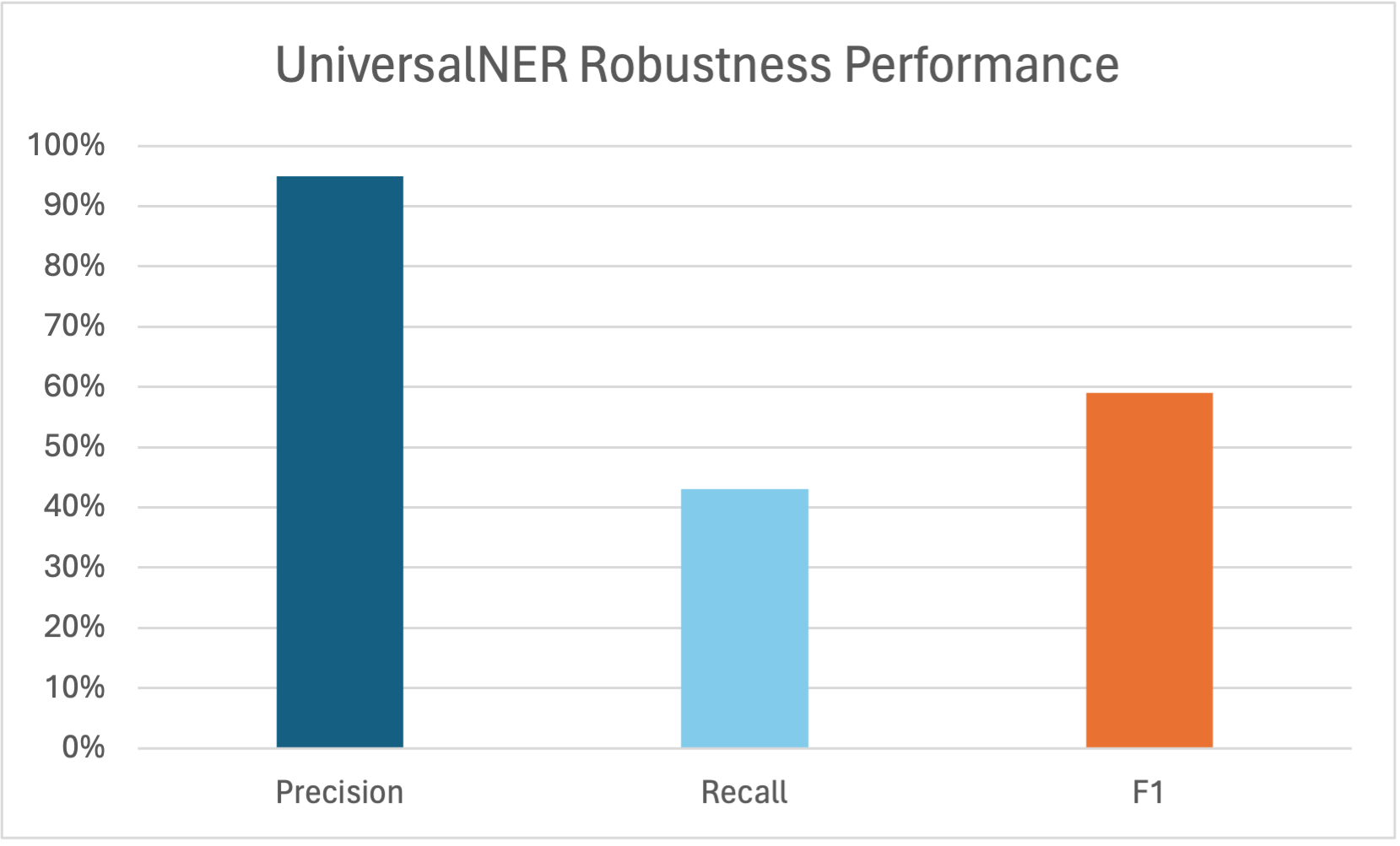}
    \caption{Robustness key performance indicators evaluation of UniversalNER-7B after fine-tuning.}
    \label{fig:robustness-evaluation}
\end{figure}


We analyzed the robustness of the fine-tuned UniversalNER model. As shown in Figure \ref{fig:robustness-evaluation}, UniversalNER achieved a Precision of 95\%, indicating accurate entity predictions with few false positives. However, the Recall was lower at 43\%, suggesting the model missed a substantial number of actual entities, leading to a high false negative rate. The F1 score of 59\% reflects this trade-off, highlighting moderate overall performance. While the model demonstrates high precision, its effectiveness is limited by its inability to identify a significant proportion of valid entities.

\section{Conclusion}
Our approach introduces three innovative information extraction approaches to enhance the activity of DNM scraping. We created our dataset containing raw HTML content of product listing pages from various DNMs. This dataset enabled us to develop three proof-of-concepts for identifying specified entities. Based on the design, analysis, and assessment of the experiments conducted, our proposed concepts effectively extract structured key-value information from raw HTML files, eliminating manual modification of information extraction scripts, as with RegEx. 

The ELMo-BiLSTM model demonstrated strong learning capabilities, with a consistent decrease in training loss across all entities, highlighting its ability to adapt to complex patterns in the data. The model showed high Precision, especially for entities like ``price'', while achieving significant improvements in Recall and F1 scores over the training epochs. However, the entities ``vendor'' and ``views'' presented initial difficulties, indicating areas where model enhancements could improve performance.

Evaluating the zero-shot scenarios, GLiNER-L performed significantly better than UniversalNER-7B on the DNMsLIsting dataset, indicating strong generalization capabilities. GLiNER-L achieved an F-score of 15\% compared to UniversalNER-7B's 6.3\%, highlighting the challenges posed by the unique characteristics of dark web data, such as significant content variations and frequent misspellings. Among the GLiNER models, GliNER-L emerged as the most effective in zero-shot learning, achieving high Precision and F1 scores, demonstrating its robust entity identification and classification abilities in unfamiliar contexts.

Fine-tuning the zero-shot models resulted in substantial performance gains, with UniversalNER-7B achieving exceptional performance (91\% Precision, 96\% recall, and an F1 score of 94\%). This highlights the effectiveness of fine-tuning specific task execution. Despite the computational efficiency of GLiNER models, UniversalNER-7B's superior performance after fine-tuning suggests that resource-intensive approaches can offer significant benefits in accuracy and robustness. 

Comparing the different proof of concept models revealed that the fine-tuned UniversalNER-7B achieved the highest overall performance, underscoring the importance of model complexity in achieving high accuracy. The ELMo-BiLSTM model, while having exact results, needed help with recall, indicating a trade-off between Precision and comprehensive entity retrieval. GLiNER-L's lower recall further remarks on the challenges in adapting to dark web data's diverse and irregular patterns.

Evaluating the robustness of UniversaNER on a different DNM (Palmetto State Armory Market) dataset demonstrated high Precision (95\%) but a lower Recall (43\%), leading to a moderate F1 score of 59\%. The performance indicates the model's predictive solid accuracy and its limitation in identifying a wide range of relevant entities in new and varied environments. Including previously unseen entities, such as ``Stock Keeping Unit Number'' and ``brand'', tests the model's adaptability and highlights areas for improvement in improving recall.

Translating our findings into concrete answers to our research questions: \\

\textit{RQ: To what extent does our proof of concept automated scraping approach for DNMs compare to state-of-the-art techniques?}

\begin{framed}
\textbf{Finding main RQ.} Our proof-of-concept approaches, particularly the fine-tuned UniversalNER-7B model, perform significantly better than state-of-the-art techniques such as ELMo-BiLSTM-CNN. The UniversalNER-7B achieved 91\% Precision, 96\% recall, and an F1 score of 94\%, compared to the ELMo-BiLSTM-CNN's 92\% Precision, 70\% recall, and 79\% F1 score. This substantial improvement, especially in recall and F1 score, indicates that our methods are highly competitive and more effective in accuracy and robustness.
\end{framed}

The improved performance could be attributed to the more extensive parameter set and comprehensive training of UniversalNER-7B, enabling it to capture a broader range of entity patterns and nuances in DNM data. Additionally, the architecture of the model plays a role in its robustness. Since UniversalNER is a pre-trained model, the number of accurate representative weights is higher than the ELMo-BiLSTM-CNN model (as can be seen from the number of parameters), leading to being more accurate.\\

\textit{sRQ1: To what extent can our proposed NER approaches scrape data from DNMs using zero-shot and fine-tuning methods?}

\begin{framed}
\textbf{Finding sRQ1.} Our NER approaches can scrape DNM data using zero-shot and fine-tuning methods. In zero-shot scenarios, GLiNER-L achieved an F-score of 15\% compared to UniversalNER-7B's 6.3\%, indicating its strong generalization. After fine-tuning process, UniversalNER-7B demonstrated exceptional performance with 91\% Precision, 96\% Recall, and an F1 score of 94\% capabilities.
\end{framed}

The substantial improvement in performance after fine-tuning underscores the effectiveness of adapting models to specific DNM datasets, allowing them to capture the unique characteristics of the data better. \\

\textit{sRQ2: Which proof of concept exhibits the best performance in automated dark web scraping with regards to recall, Precision, and F-1} 

\begin{framed}
\textbf{Finding sRQ2.}
The fine-tuned UniversalNER-7B model performs best in automated dark web scraping, achieving 91\% Precision, 96\% recall, and an F1 score of 94\%.
\end{framed}

The ELMO-BiLSTM-CNN model achieved 92\% Precision but only 70\% recall and an F1 score of 79\%. The GliNER-L model was the best-performing GLiNER. After fine-tuning, GLiNER-L reached 88\% Precision, 59\% recall, and an F1 score of 71\%. UniversalNER-7B's superior performance underscores its effectiveness and adaptability in extracting entities from DNM data, demonstrating its ability to balance high Precision with comprehensive entity retrieval. The performance of the UniversalNER made it the best-performing model among the models treated in this study. \\

\textit{sRQ3: In comparison with existing studies, how robust are the proposed approaches for automated dark web scraping?}

\begin{framed}
\textbf{Finding sRQ3.}
UniversalNER-7b model demonstrates significant robustness. However, the irregular patterns of DNM point to the need for more varied data to enhance adaptability and improve recall in dark web scraping tasks. 
\end{framed}

As stated, robustness is defined as the method's ability to maintain high performance across diverse web environments and adapt to future web technologies and standards. The fine-tuned UniversalNER-7b model demonstrates significant robustness, maintaining high Precision (95\%) across different datasets, including those with previously unseen entities. However, its recall is lower (43\%) in new environments, leading to an F1 score of 59\%. These findings indicate the model's predictive solid accuracy but highlight areas where further improvements are needed to enhance robustness and adaptability. The ability to handle diverse and irregular patterns in DNM data underscores the importance of fine-tuning to achieve higher robustness in entity extraction tasks. More DNM data is needed to enrich the model and improve the recall of the model and, therefore, the F1 score. 

To conclude,  our work proposed various approaches for information extraction from dark web data. While zero-shot approaches offer valuable insights and robust initial capabilities, fine-tuning enhances accuracy and recall significantly. The UniversalNER-7B model, despite its resource-intensive nature, proved to be the most effective in this context, demonstrating the potential of advanced NER techniques in handling the challenges of dark web scraping.

\section{Discussion and Threat to Validity}
To evaluate the validity of our study, we employ the framework of threats to validity (TTV), which identifies factors that may compromise the reliability and validity of our findings. The TTV assessment serves two primary purposes: first, to contextualize the results and identify potential influences on the conclusions, and second, to outline mitigation strategies, including how threats were addressed and why certain limitations remain \citep{VerdecchiaEtAl2023}. TTV is typically categorized into internal and external validity. \textbf{Internal validity} refers to the extent to which the results can be attributed to the variables intentionally manipulated or measured rather than being influenced by extraneous factors or biases. It assesses the accuracy and correctness of the study within its controlled environment. Conversely, \textbf{external validity} concerns the generalizability of the findings to different settings, populations, and times beyond the specific conditions of the original research. It evaluates how applicable and transferable the results are to real-world scenarios outside the controlled environment \citep{SiegmundEtAl2015}.

Several factors impact the \textbf{internal validity} of this study. A critical component is the dataset, which forms the foundation for the developed models. The data was systematically collected from various darknet markets (DNMs), with a significant portion sourced from the Cocorico marketplace. While this approach expands the dataset and mitigates risks of model overfitting or underfitting, it may also introduce sampling bias. Nonetheless, preprocessing and normalization steps, including parsing HTML pages and eliminating noise and irrelevant tags, ensure consistency in the text data used for training and evaluation. This consistency reduces variability that could affect model performance.

Training procedures were rigorously defined, with uniform hyperparameters and model architectures across experiments. Evaluation metrics (Precision, Recall, and F1-score) were consistently applied, ensuring comparability of results. Additionally, the robustness of the models was tested on a different DNM dataset, demonstrating their ability to generalize beyond the initial dataset.

Potential threats to internal validity include the reliance on RegEx patterns for labeling, which may introduce bias if the patterns fail to capture all data variations. This risk was mitigated through manual verification, maintaining an accuracy threshold of over 90\% for RegEx patterns. However, using different hyperparameters poses a threat, as alternative settings could yield different outcomes, suggesting a need for further exploration. Another concern is the use of smaller validation datasets for the GLiNER model due to out-of-memory (OOM) issues, which could affect the comparability of results.



The \textbf{external validity} of this study is reinforced by several factors that enhance the generalizability of its findings. The dataset includes product listings from six DNMs with diverse structures and content, spanning English and French. This diversity suggests that the approaches could be applicable to other DNMs in similar contexts. Moreover, the use of zero-shot approaches, such as UniversalNER and GLiNER, followed by fine-tuning, demonstrates the adaptability of these models to new datasets, indicating their potential applicability to other named entity recognition (NER) tasks in different domains with appropriate modifications. Robustness evaluations on the Palmetto State Armory Market dataset further support these findings.

However, the study's focus on DNMs, which have unique characteristics such as specific jargon and webpage structures, may limit the generalizability of the findings to non-DNM contexts without further adaptation. Additionally, while the dataset includes English and French, it does not account for other languages, leaving the performance of the models on non-English DNM data untested. Lastly, the techniques and tools employed, such as ELMo embeddings and LoRA for fine-tuning, may require modifications to be directly applicable to other NER tasks.

Finally, it is important to note that various other information extraction approaches were tested, including a Scrapy NER model and a BERT-embedded BiLSTM model. However, these methods were excluded from the final analysis due to their limitations and suboptimal performance. For instance, the Scrapy NER model's reliance on hard-coded RegEx patterns rendered it non-robust and unsuitable for automated information extraction in this context.

\section{Limitations}
Zooming in on the threats to validity, a significant limitation is the high computational resources required for training and fine-tuning the UniversalNER and GLiNER language models. Despite employing memory-efficient techniques such as LoRA, the fine-tuning process remains resource-intensive. These constraints limited the extent of hyperparameter tuning and robustness testing for the GLiNER model. The substantial resource demands also pose challenges for law enforcement agencies (LEAs) with limited access to advanced hardware, such as high-capacity GPUs. This reliance on resource-heavy computational power not only escalates costs but also restricts the models' accessibility and scalability.

Another limitation is the linguistic diversity within the DNMsListings dataset, which currently includes only English and French content. This narrow linguistic scope limits the models' generalizability to other languages used on DNMs, potentially leading to poor performance on datasets featuring languages with different grammatical structures, vocabularies, or usage patterns.

More broadly, the availability of dark web data is a constraint. The dataset's size influences the robustness and generalizability of the findings. Although the dataset contains HTML documents from six different DNMs, expanding this collection could enhance model recall and overall performance. Additionally, model robustness was tested on data from only one additional DNM, which may not fully capture the diversity and variability across different DNMs. Increasing the amount of dark web data could positively impact the findings and improve the models' generalizability.

\section{Future Work}



Future research should focus on enhancing model efficiency and scalability by exploring lighter architectures or resource-efficient fine-tuning methods. Expanding datasets from diverse darknet markets (DNMs) is also key to improving model robustness and generalizability. Additionally, integrating these models into dark web crawlers and addressing anti-scraping mechanisms would improve their applicability for real-time insights and law enforcement use.

\bibliographystyle{cas-model2-names}
\bibliography{bibliography}





\end{document}